\pdfoutput=1
\documentclass[leqno,onefignum,onetabnum]{siamltex}
\usepackage[left=2.5cm,right=2.5cm, top=2cm, bottom=2cm]{geometry}

\usepackage{fullpage}

\usepackage{microtype}
\usepackage{graphicx}
\usepackage{caption}
\usepackage{subcaption}
\usepackage{booktabs} 
\usepackage{bm}
\newcommand{\nexp}{\text{\sc{e}-}}
\newcommand{\nexpb}{\text{\sc{\textbf{e}}-}}
\newcommand{\pexp}{\text{\sc{e}{\scriptsize +}}}

\usepackage[square,sort&compress,comma,numbers]{natbib}

\usepackage[utf8]{inputenc} 
\usepackage[T1]{fontenc}    
\usepackage{url}            
\usepackage{booktabs}       
\usepackage{amsfonts}       
\usepackage{nicefrac}       
\usepackage{microtype}      
\usepackage{soul}
\usepackage[utf8]{inputenc}
\usepackage{amsmath}
\usepackage{booktabs}
\usepackage{outlines}
\usepackage{adjustbox}

\usepackage{chngcntr}

\usepackage[colorlinks=true,
citecolor=blue,
filecolor=black,
linkcolor=blue,
urlcolor=blue]{hyperref}

\usepackage{microtype}
\usepackage{graphicx}
\usepackage{booktabs} 

\usepackage{multirow}
\usepackage{paralist}

\usepackage{booktabs} 
\usepackage{multicol}
\usepackage{diagbox}

\usepackage{here}

\usepackage{amsmath,amssymb,amsfonts,amsbsy,amsfonts,latexsym}
\usepackage{multirow}
\usepackage{makecell}
\usepackage[labelfont=bf,textfont=it,belowskip=0pt,aboveskip=5pt,tableposition=top]{caption}
\usepackage{xcolor}
\usepackage{colortbl}

\definecolor{colorA}{RGB}{189,201,225}
\definecolor{colorB}{RGB}{103,169,207}
\definecolor{colorC}{RGB}{ 28,144,153}
\definecolor{colorD}{RGB}{  1,108, 89}

\usepackage{tabularx,colortbl,xcolor}
\usepackage{multirow}
\usepackage[normalem]{ulem}
\useunder{\uline}{\ul}{}

\usepackage{enumitem}

\usepackage{xparse}

\DeclareGraphicsExtensions{.pdf,.png}

\usepackage{longtable}
\usepackage{pgfplots}

\usepackage{lipsum}

\usepackage{amssymb}
\usepackage{pifont}

\newcommand\sref{\S\ref}
\newcommand\aref{Algorithm~\ref}
\newcommand\eref{Eq.~\ref}
\newcommand\fref{Fig.~\ref}
\newcommand\tref{Tab.~\ref}

\usepackage{adjustbox}

\newcommand\gd{ \rowcolor{gray!45}}

\newcolumntype{L}{>{\columncolor{gray!20}}c}

\usepackage{wrapfig}

\usepackage{amsmath}

\usepackage{algorithm}
\usepackage{algpseudocode}

\newcommand{\idiv}{\ensuremath{\mbox{div}\,}}
\newcommand{\igrad}{\ensuremath{\nabla}}

\newcommand{\vect}[1]{\boldsymbol{#1}} 
\newcommand{\mat}[1]{\boldsymbol{#1}}  
\newcommand{\resampling}{\textsc{Resampling}}
\newcommand{\adaptiver}{\textsc{Adaptive-R}}
\newcommand{\adaptiveg}{\textsc{Adaptive-G}}
\newcommand{\adaptiverns}{\textsc{Adaptive-RNS}}
\newcommand{\adaptivegns}{\textsc{Adaptive-GNS}}

\newcommand{\TCone}{\textsc{TC1}}
\newcommand{\TCtwo}{\textsc{TC2}}
\newcommand{\TCthr}{\textsc{TC3}}
\newcommand{\TCfou}{\textsc{TC4}}

\usepackage{fancyhdr}
\title{Adaptive Self-supervision Algorithms for Physics-informed Neural Networks}

\author{
Shashank Subramanian$^{1}$, 
Robert M. Kirby$^{2}$, 
Michael W. Mahoney$^{1,3,4}$, 
Amir Gholami$^{4}$   \\
$^{1}$Lawrence Berkeley National Laboratory, 
$^{2}$University of Utah,
$^{3}$International Computer Science Institute, \\
$^{4}$University of California, Berkeley   \\
{\tt\small shashanksubramanian@lbl.gov, kirby@cs.utah.edu, \{mahoneymw, amirgh\}@berkeley.edu,\ }
}

\begin{document}

\maketitle
\thispagestyle{empty}

\setcounter{page}{1}

\interfootnotelinepenalty=10000

\begin{abstract}
Physics-informed neural networks (PINNs) incorporate physical knowledge from the problem domain as a soft constraint on the loss function, but recent work has shown that this can lead to optimization difficulties. 
Here, we study the impact of the location of the collocation points on the trainability of these models. 
We find that the vanilla PINN performance can be significantly boosted by adapting the location of the collocation points as training proceeds. 
Specifically, we propose a novel adaptive collocation scheme which progressively allocates more collocation points (without increasing their number) to areas where the model is making higher errors (based on the gradient of the loss function in the domain).  
This, coupled with a judicious restarting of the training during any optimization stalls (by simply resampling the collocation points in order to adjust the loss landscape) leads to better estimates for the prediction error.
We present results for several problems, including a 2D Poisson and diffusion-advection system with different forcing functions. 
We find that training vanilla PINNs for these problems can result in up to 70\% prediction error in the solution, especially in the regime of low collocation points. 
In contrast, our adaptive schemes can achieve up to an order of magnitude smaller error, with similar computational complexity as the baseline. 
Furthermore, we find that the adaptive methods consistently perform on-par or slightly better than vanilla PINN method, even for large collocation point regimes.
The code for all the experiments has been open sourced and available at~\cite{github}.
\end{abstract}
\section{Introduction}\label{sec:intro}
A key aspect that distinguishes scientific ML (SciML) \cite{karniadakis2021review, von2019informed, willard2020integrating, brunton2020machine,rackauckas2020universal, lu2021deepxde} from other ML tasks is that scientists typically know a great deal about the underlying physical processes that generate their data.
For example, while the Ordinary Differential Equations (ODEs) or Partial Differential Equations (PDEs) used to simulate the physical phenomena may not capture every detail of a physical system, they often provide a reasonably good approximation.
In some cases, we know that physical systems have to obey conservation laws (mass, energy, momentum, etc.).
In other cases, we can learn these constraints, either exactly or approximately, from the data.
In either case, \textit{the main challenge in SciML lies in combining such scientific prior domain-driven knowledge with large-scale data-driven methods from ML in a principled~manner}.

One popular method to incorporate scientific prior knowledge is to incorporate them as soft-constraints
throughout training, as proposed with Physics Informed Neural Networks (PINNs)~\cite{raissi2019physics,karniadakis2021review, lagaris1998artificial}. 
These models use penalty method techniques from optimization~\cite{boyd2004convex} 
and formulate the solution of the PDE as an unconstrained optimization problem that minimizes a self-supervision loss function that incorporates the domain physics (PDEs) as a penalty (regularization) term.
Formulating the problem as a soft-constraint makes it very easy to use existing auto-differentiation frameworks
for SciML tasks. 
However, training PINNs this way can be very difficult, and it is often difficult to 
solve the optimization problem~\cite{krishnapriyan2021characterizing,wang2020understanding,EdwCACM22}.
This could be partly because the self-supervision term typically contains complex terms such as (higher-order) derivatives of spatial functions and other nonlinearities that cause the loss term to become ill-conditioned \cite{krishnapriyan2021characterizing}. 
This is very different than unit $\ell_p$ ball or other such convex functions, more commonly used as regularization terms in ML.
Several solutions such as loss scaling~\cite{wang2020understanding}, curriculum or sequence-to-sequence learning~\cite{krishnapriyan2021characterizing}, tuning of loss function weights~\cite{liu2021dual}, and novel network architectures~\cite{ramabathiran2021spinn} have been proposed to address this~problem.

One overlooked, but very important, parameter of the training process is the way that the self-supervision is performed in PINNs,
and in particular which data points in the domain are used for enforcing the physical constraints (commonly referred to as \emph{collocation points} in numerical analysis).
In the original work of~\cite{raissi2019physics}, the collocation points are randomly sampled in the beginning of the
training and kept constant throughout the learning process. However, we find that this is sub-optimal,
and instead we propose adaptive collocation schemes.
We show that these schemes can result in an order of magnitude better performance, with
similar computational overhead.

\paragraph{Background}
We focus on scientific systems that have a PDE constraint of the following form:
\begin{equation}
     \mathcal{F} (u(\vect{x})) = 0, \qquad \vect{x} \in \Omega \subset \mathbb{R}^{d}, 
    \label{eq:pde_problem_statement}
\end{equation}
where $\mathcal{F}$ is a differential operator representing the PDE, $u(\vect{x})$ is the state variable (i.e., physical quantity of interest), $\Omega$ is the physical domain, and $\vect{x}$ represents spatial domain (2D in all of our results). To ensure existence and uniqueness of an analytical solution, there are additional constraints specified on the boundary, $d\Omega$, as well (such as periodic or Dirichlet boundary conditions).
One possible approach to learn a representation for the solution is to incorporate the PDE
as a hard constraint, and formulate a loss function that measures the prediction error on the boundary
(where data points are available):
\begin{equation}
    \min_{\theta} \mathcal{L}(u) \quad\text{s.t.}\quad \mathcal{F}(u) = 0,
\label{eq:pinns_optimization_problem_gen}
\end{equation}
where $\mathcal{L}(u)$ is typically a data mismatch term (this includes initial/boundary conditions but can also include observational data points), and where $\mathcal{F}$ is a constraint on the residual of the PDE system (i.e., $\mathcal{F}(u)$ is the residual) under consideration.
Since constrained optimization is typically more difficult than unconstrained optimization~\cite{boyd2004convex}, this constraint is typically relaxed and added as a penalty term to the loss function.
This yields the following unconstrained optimization problem, namely the PINNs (soft constrained) optimization problem:

\begin{equation}
\min_{\theta} \mathcal{L}(u) + \lambda_{\mathcal{F}} \mathcal{L}_\mathcal{F}. 
 \label{eq:pinns_optimization_problem_naive}
\end{equation}
In this problem formulation, 
the regularization parameter, $\lambda_{\mathcal{F}}$, controls the weight given to the PDE constraints, as compared to the data misfit term;
$\theta$ denotes the parameters of the model that predicts $u(\vect{x})$, which is 
often taken to be a neural network; and the PDE loss functional term, $\mathcal{L}_\mathcal{F}$, can be considered as a self-supervised loss, as all the information comes from the PDE system we are interested in simulating, instead of using direct observations.
Typically a Euclidean loss function is used to measure the residual of the PDE for this loss function, $\|\mathcal{F}(u)\|_2^2$, where the $\ell_2$ norm is computed at discrete points (collocation points) that are randomly sampled from $\Omega$. This loss term is often the source of the training difficulty with PINNs~\cite{krishnapriyan2021characterizing,EdwCACM22}, which is the focus of our paper.

\paragraph{Main contributions}
Unlike other work in the literature, which has focused on changing the training method or the neural network (NN) architecture, here we
focus on the self-supervision component of PINNs, and specifically on the selection of the collocation points. 
In particular, we make the following contributions:

\begin{itemize}
    \item We study the role of the collocation points for two PDE systems: steady state diffusion (Poisson); and diffusion-advection. We find that keeping the collocation points constant
    throughout training is a sub-optimal strategy and is an important source of the training difficulty with PINNs.
    This is particularly true for cases where the PDE problem exhibits local behaviour (e.g., in presence of sharp, or very localized, features).
    
    \item We propose an alternative strategy of resampling the collocation points when training stalls.  Although this strategy is simple, it can lead to significantly
    better reconstruction (see~\tref{tab:poisson} and~\fref{fig:poisson}). Importantly, this approach
    does not increase the computational complexity, and it is easy to~implement.
    
    \item We propose to improve the basic resampling scheme with a gradient-based adaptive scheme.  This adaptive scheme is designed to help to relocate
    the collocation points to areas with higher loss gradient, without increasing the total number of points (see~\aref{alg:adapt} for the algorithm). In particular, we progressively relocate the points to areas of high gradient
    as training proceeds. This is done through a cosine-annealing that gradually changes the sampling of collocation points from uniform to adaptive through training. 
    We find that this scheme consistently achieves better performance than the basic resampling method, and it can lead
    to more than 10x improvements in the prediction error (see~\tref{tab:poisson}, ~\fref{fig:poisson}).
    
    \item We extensively test our adaptive schemes for the two PDE systems of Poisson and diffusion-advection while
    varying the number of collocation points for problems with both smooth or sharp features.
    While the resampling and adaptive schemes perform similarly in the large collocation point regime, the adaptive approach shows significant improvement when the number of collocation points is small and the forcing function is sharp (see~\tref{tab:poisson_col}).
 \end{itemize}

\section{Related work}
\label{sec:related_work}

There has been a large body of work studying PINNs~\cite{raissi2020hidden, chen2020physics, jin2021nsfnets, sirignano2018dgm, zhu2019physics, geneva2020modeling, sahli2020physics} and the challenges associated with
their training~\cite{EdwCACM22,wang2020and, wang2020eigenvector,wang2020understanding,krishnapriyan2021characterizing,wang2022respecting}. The work of~\cite{wang2020understanding} notes these challenges and proposes a loss scaling
method to resolve the training difficulty. 
Similar to this approach, some works have treated the problem
 as a multi-objective optimization and tune the weights of the different loss terms \cite{xiang2021self,bischof2021multi}.
A more formal approach was suggested in \cite{liu2021dual} where the weights are learned by solving a minimax optimization problem that ascends in the loss weight space and descends in the model parameter space. This approach was extended in \cite{mcclenny2020self} to shift the focus of the weights from the loss terms to the training data points instead, and the minimax forces the optimization to pay attention to specific regions of the domain. 
However, minimax optimization problems are known to be hard to optimize and introduce additional complexity and computational expense. 
Furthermore, it has been shown that using
curriculum or sequence-to-sequence learning can ameliorate the training difficulty with PINNs~\cite{krishnapriyan2021characterizing}. More recently, the work of~\cite{wang2022respecting} shows
that incorporating causality in time can help training for time-dependent PDEs.

There is also recent work that studies the role of the collocation points. For instance,~\cite{tang2021deep} refines the collocation point set without learnable weights.
They propose an auxiliary NN that acts as a generative model to sample new collocation points that mimic the PDE residual.
However, the auxiliary network also has to be trained in tandem with the PINN. 
The work of~\cite{lu2021deepxde} proposes an adaptive collocation scheme where
the points are densely sampled uniformly and trained for some number of iterations. Then the set is extended by adding points in increasing rank order of PDE residuals to refine in certain locations (of sharp fronts, for example) and the model is retrained. However, this method can increase the computational overhead, as the number of collocation points
is progressively increased.
Furthermore, in
\cite{hanna2021residual} the authors 
show that the latter approach can lead to excessive clustering of points throughout training.
To address this, instead they propose to add points based on an underlying density function defined by the PDE residual. Both these schemes keep the original collocation set (the low residual points) and increase the training dataset sizes as the optimization proceeds. 

Unlike the work of~\cite{lu2021deepxde, hanna2021residual}, we focus on using gradient of the loss function, instead
of the nominal loss value, as the proxy to guide the adaptive resampling of the collocation points. We show that this
approach leads to better localization of the collocation points, especially for problems with sharp features.
Furthermore, we incorporate a novel cosine-annealing scheme, which progressively incorporates adaptive sampling as
training proceeds. Importantly, we also keep the number of collocation points the same.
Not only does this not increase the computational overhead,
but this is also easier to implement as well.

\section{Methods}
In PINNs, we use a feedforward NN, denoted $\textit{NN}(\vect{x};\theta)$, that is parameterized by weights and biases, $\theta$, takes as input values for coordinate points, $\vect{x}$, and outputs the solution value $u(\vect{x}) \in \mathbb{R}$ at these points.
As described in \sref{sec:intro}, the model parameters $\theta$ are optimized through the loss function:
\begin{equation}
\min_{\theta} \mathcal{L}_\mathcal{B} + \lambda_{\mathcal{F}} \mathcal{L}_\mathcal{F}.
 \label{eq:pinns_optimization_problem}
\end{equation}
We focus on boundary-value (steady state) problems and define the two loss terms as:
\begin{subequations}
    \begin{equation}
        \mathcal{L}_\mathcal{B} = \frac{1}{n_b}\sum_{i=1}^{n_b} \|u(\vect{x}_b^i) - \hat{u}(\vect{x}_b^i)\|_2^2, \label{eq:loss_bc}
    \end{equation}
    \begin{equation}
        \mathcal{L}_\mathcal{F} = \frac{1}{n_c}\sum_{i=1}^{n_c} \|\mathcal{F}(u(\vect{x}_c^i)\|_2^2,
    \label{eq:loss_pde}
    \end{equation}
\end{subequations}
where $u$ is the model predicted solution, $\hat{u}$ is the true solution or data, $\vect{x}_b^i$ are points on the boundary, and $\vect{x}_c^i$ are collocation points uniformly sampled from the domain $\Omega$. 
Here, $n_b$ and $n_c$ are the number of boundary and collocation points, respectively; and the boundary loss term $\mathcal{L}_\mathcal{B}$ implements a Dirichlet boundary condition, where we assume that the solution values are known on the boundary $d\Omega$.
In PINNs~\cite{raissi2019physics}, the collocation points used in~\eref{eq:loss_pde} are randomly sampled with a uniform probability over the entire space $\Omega$
in the beginning of training and then kept constant afterwards (we refer to this
approach as Baseline).
While a uniformly distributed collocation point set may be sufficient for simple PDEs with smooth features, we find them to be sub-optimal when the problem exhibits sharp/local features, or even fail to train. To address this, we propose the following
schemes.

\paragraph{Resampling collocation points}
Current PINNs are typically optimized with LBFGS.
In our experiments, we found that in the baseline approach LBFGS often fails to find
a descent direction and training stalls, even after hyperparamter
tuning.
This agrees with other results reported in the literature~\cite{wang2020understanding, krishnapriyan2021characterizing,EdwCACM22}.
We find that this is partially due to the fact that the collocation points
are kept constant and not changed. The simplest approach to address this
is to resample the collocation points when LBFGS stalls. We refer to this approach
as \resampling{}. 
As we will discuss in the next section, we find this approach to be helpful for cases with moderate to large number of collocation points.

\paragraph{Adaptive sampling} 
While the \resampling{} method is effective for large number of collocation points,
we found it to be sub-optimal in the small collocation point regime, especially for problems
with sharp/localized features. In this case, the \resampling{} method still uses a uniform distribution
with which to sample the new the collocation points; and, in the presence of a small number of collocation points and/or sharp/localized features, this is not an optimal allocation of the points.
Ideally, we want to find a probability distribution, as a replacement for the
uniform distribution, that can improve trainability of PINNs for
a given number of collocation points and/or computational budget.
There are several possibilities to define this distribution.
The first intuitive approach would be to use the value of the PDE residual 
(\eref{eq:loss_pde}), and normalize
it as a probability distribution. This could then be used to
sample collocation points based on the loss values
in the domain. That is, more points would be sampled in areas with 
with higher PDE residual, and vice versa.
We refer to this approach as \adaptiver{} sampling
(with $\textsc{R}$ referring to the PDE residual).
An alternative is to use the gradient of the loss function (PDE loss term $\mathcal{L}_\mathcal{F}$) w.r.t. the 
spatial grid, using that as the probability distribution with which to sample
the collocation points. We refer to this approach as \adaptiveg{} (with $\textsc{G}$ referring to gradient of the loss).
As we
will show in the next section, when combined with a cosine annealing scheme
(discussed next), both of these adaptive schemes consistently
perform better than the \resampling{} scheme.

\paragraph{Progressive adaptive sampling} 
Given the non-convex nature of the problem, it might be important to not overly constrain the NN to be too locally focused, as
the optimization procedure could get stuck in a local minima. However, this can
be addressed by progressively incorporating adaptive sampling as training proceeds.
In particular, we use a cosine annealing strategy. This allows the NN to
periodically alternate between focusing on regions of high error as well as uniformly covering larger parts of the sample space, over
a period of iterations.
Specifically, in each annealing period, we start with a full uniform sampling, and progressively incorporate adaptively
sampled points using a cosine schedule rule of $\eta = 1/2(1 + \cos{\pi T_c/T})$,
where $\eta$ is the fraction of points uniformly sampled, $T_c$ is the number of epochs since the last restart, and $T$ is the length of the cosine schedule.
We use this schedule to resample the collocation every $e$ epochs. Given the periodic nature of the cosine annealing,
that approach balances local adaptivity without losing global information.
We outline the adaptive sampling \aref{alg:adapt}. 

\begin{algorithm}
\caption{Adaptive Sampling for Self-supervision in PINNs}\label{alg:adapt}
\begin{algorithmic}[1]
\Require Loss $\mathcal{L}$, NN model, number of collocation points $n_c$, PDE regularization $\lambda_\mathcal{F}$, $T$, $s_w$, momentum $\gamma$, max epochs $i_\text{max}$
\State $i \gets 0$
\While{$i \leq i_\text{max}$}
    \State Compute proxy function as the loss gradient (\adaptiveg{}) or PDE residual (\adaptiver{)}
    \State Current proxy $\mathcal{P}_i \gets \mathcal{P} + \gamma \mathcal{P}_{i-1}$
    \State $T_c \gets i ~\text{mod} ~ T$ 
    \State $\eta \gets \text{cosine-schedule}(T_c, T)$ 
    \If{$i ~ \text{mod} ~ e$ is true}
    \State Sample $\eta n_c$ points $\vect{x}_u$ uniformly
    \State Sample $(1 - \eta) n_c$ points $\vect{x}_a$ using proxy function $\mathcal{P}_i$
    \EndIf
    \State $\vect{x}_c \gets \vect{x}_u \cup \vect{x}_a$
    \State Input $\vect{x} \gets \vect{x}_b \cup \vect{x}_c$ where $\vect{x}_b$ are boundary points
    \State $u \gets \textit{NN}(\vect{x};\theta)$
    \State $\mathcal{L} \gets \mathcal{L}_\mathcal{B} + \lambda_\mathcal{F} \mathcal{L}_\mathcal{F}$
    \State $\theta \gets \text{optimizer-update}(\theta, \mathcal{L})$
    \If{stopping-criterion is true}
        \State reset cosine scheduler $T_c \gets 0$
    \EndIf
    \State $i \gets i+1$
\EndWhile
\end{algorithmic}
\end{algorithm}

\section{Experiments}
In this section, we show examples that highlight the prediction error improvement using our adaptive self-supervision method on two PDE systems: Poisson's equation \sref{sec:poisson} and (steady state) diffusion-advection \sref{sec:poisadv}. 

\paragraph{Problem setup}
We use the same problem set up as in~\cite{raissi2019physics,wang2020understanding,krishnapriyan2021characterizing} for the NN model, which is a feed forward model with hyperbolic tangent activation function, trained
with LBFGS optimizer.
We focus on 2D spatial domains in $\Omega = [0,1]^2$.
The training data set contains points that are randomly sampled from a uniform $256^2$ mesh on $\Omega$. The testing data set is the set of all $256^2$ points, along with the true solution of the PDE $\hat{u}(\vect{x}$) at these points. Furthermore, we use a constant
regularization parameter of $\lambda_\mathcal{F}=1\nexp{4}$ for all the runs,
which we found to be a good balance between the boundary term and the PDE residual.
To ensure a fair comparison, we tune the rest of the hyperparameters both for the baseline model
as well as for the adaptive schemes. For tuning the parameters, we use the validation
loss computed by calculating the total loss over randomly sampled points in the domain
(30K points for all the experiments). Note that we do not use any signal from
the analytical solution, and instead only use the loss in~\eref{eq:pinns_optimization_problem}.

\paragraph{Metrics} 
We train the optimizer for $5000$ epochs with a stall criterion when the loss does not change for $10$ epochs. We keep track of the minimum validation loss in order to get the final model parameters for testing. We compute our prediction error using two metrics: the $\ell_2$ relative error, $\mu_1 = \|u - \hat{u}\|_2/\|\hat{u}\|$ and the $\ell_1$ error $\mu_2 = \|u - \hat{u}\|_1$, where $u$ is the model prediction solution from the best validation loss epoch and $\hat{u}$ is the true solution.
All runs are trained on an A100 NVIDIA GPU on the Perlmutter supercomputer.

\vspace{8mm}
\subsection{2D Poisson's equation}\label{sec:poisson}
\begin{figure}
\begin{subfigure}{0.5\textwidth}
  \centering
  \includegraphics[width=\linewidth]{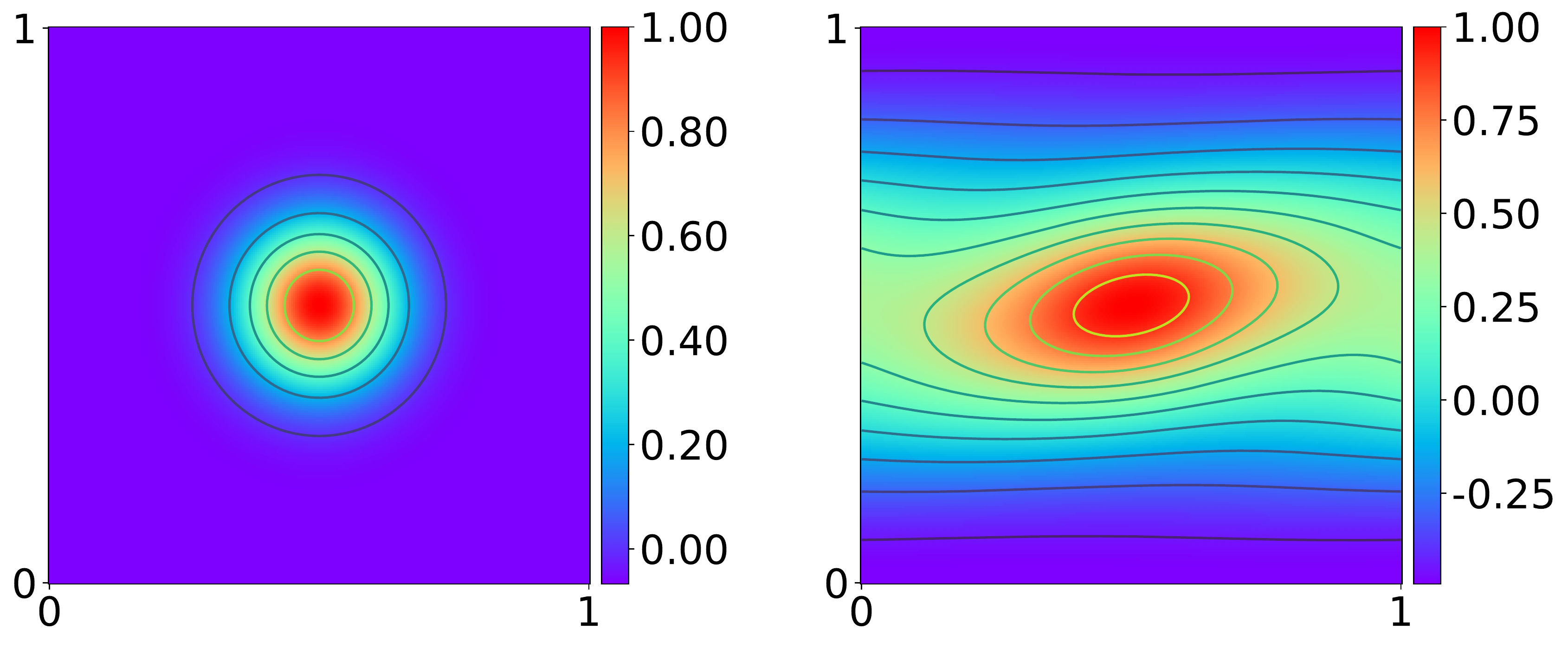}  
  \caption{$TC1$}
  \label{fig:poisson-smooth}
\end{subfigure}%
\begin{subfigure}{0.5\textwidth}
  \centering
  \includegraphics[width=\linewidth]{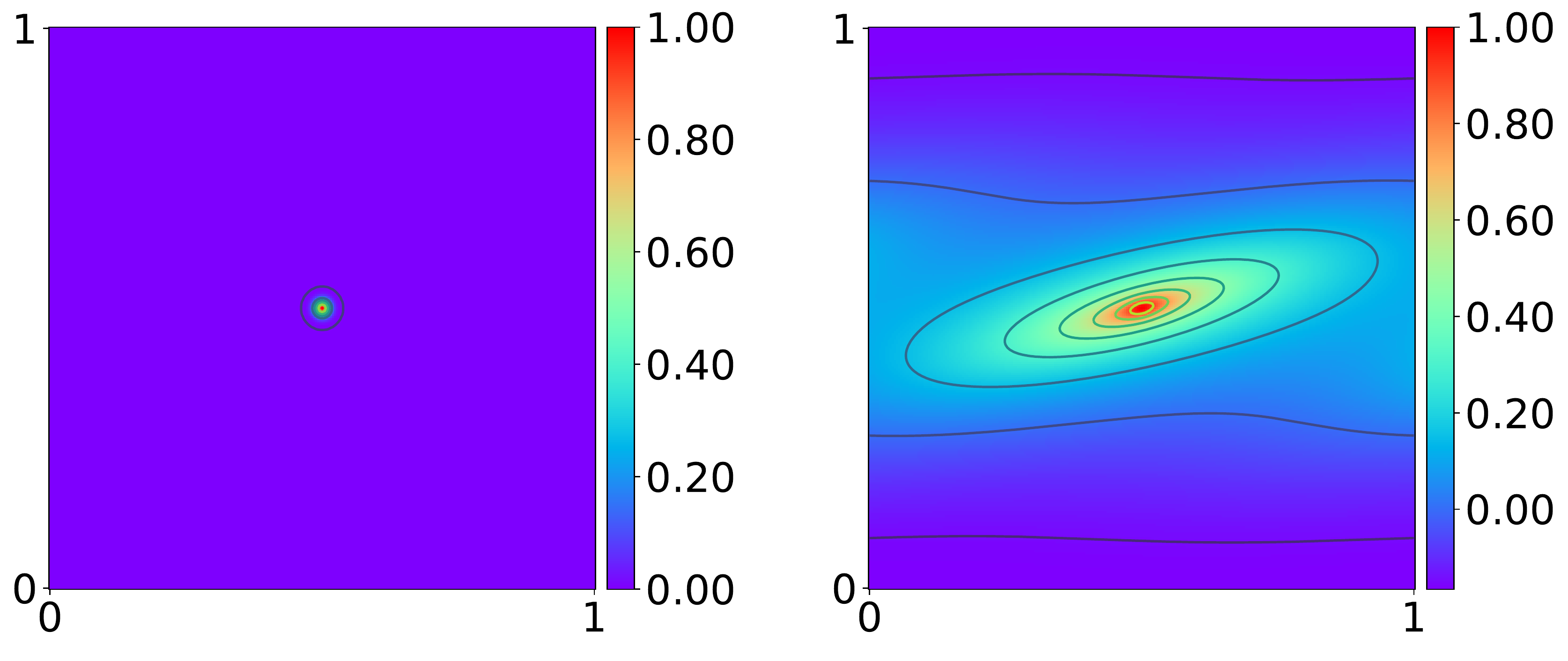}  
  \caption{$TC2$}
  \label{fig:poisson-sharp}
\end{subfigure}
    \caption{We visualize the source and target solution pairs for \TCone{} (smooth source with $\sigma_f=0.1$, left) and \TCtwo{} (sharp source with $\sigma_f=0.01$, right) for the 2D Poisson's equation.}
    \label{fig:poisson_systems}
\end{figure}

\paragraph{Problem formulation}
We consider a prototypical elliptic system defined by the Poisson's equation with source function $f(\vect{x})$. This system represents a steady state diffusion equation:
\begin{equation}
    -\idiv \mat{K} \igrad u = f(\vect{x}), \quad \vect{x} \in \Omega,
    \label{eq:poisson}
\end{equation}
where $\mat{K}$ denotes the diffusion tensor. For homogeneous diffusion tensors, the solution of the poisson's equation with doubly-periodic boundary conditions can be computed using the fast Fourier transform on a discrete grid as:
\begin{equation}
    \hat{u} = F^{-1}\big(\frac{-1}{-(k_x^2k_{11} + k_y^2k_{22} + 2k_xk_yk_{12})} F(f(\vect{x})) \big),
\end{equation}
where $F$ is the Fourier transform, $k_x, k_y$ are the frequencies in the Fourier domain, and $k_{11}, k_{22}, k_{12}$ are the diagonal and off-diagonal coefficients of the diffusion tensor $\mat{K}$.
We enforce the boundary conditions as Dirichlet boundary conditions using the true solution to circumvent the ill-posedness of the doubly-periodic Poisson's equation.\footnote{Note that the NN reaches suboptimal solutions with periodic boundaries due to the forward problem ill-posedness.} 
We use a Gaussian function as the source with standard deviation $\sigma_f$ and consider the diffusion tensor $k_{11}=1, k_{22}=8, k_{12}=4$ to make the problem anisotropic. We consider two test-cases:
\begin{inparaenum}[(i)]
\item \TCone{}: smooth source function with $\sigma_f = 0.1$
\item \TCtwo{}: sharp source function with $\sigma_f = 0.01$.
\end{inparaenum}
We show these source functions and the corresponding target solutions in \fref{fig:poisson_systems}. For any experiment, we sweep over all the hyperparameters and select the ones that show the lowest/best validation loss for the final model. 

\begin{figure}
\begin{subfigure}{\textwidth}
  \centering
  \includegraphics[width=\linewidth]{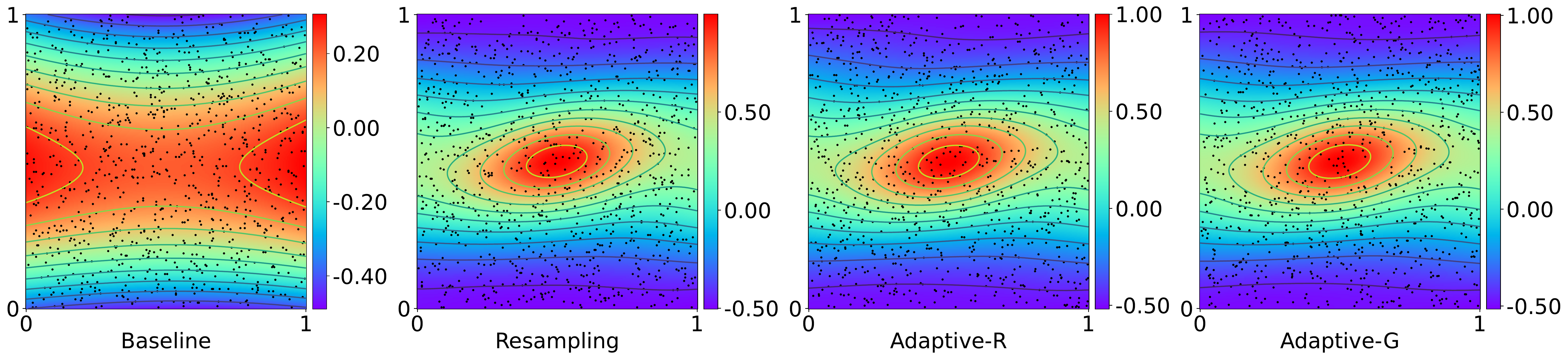}  
  \caption{$TC1$}
  \label{fig:poisson-smooth-vis}
\end{subfigure}\\
\begin{subfigure}{\textwidth}
  \centering
  \includegraphics[width=\linewidth]{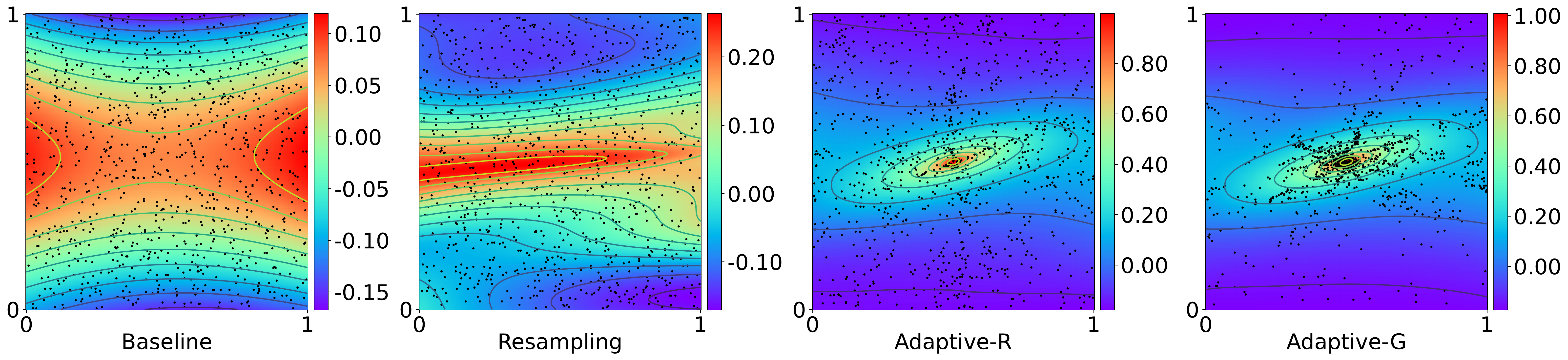}  
  \caption{$TC2$}
  \label{fig:poisson-sharp-vis}
\end{subfigure}
\caption{We visualize the predicted solution at all testing points for both test-cases TC1 (top) and TC2 (bottom). We also visualize the location of the collocation points for the final model. We observe that for the smooth source, the resampling helps predict a good solution, while the sharp source functions still cause the model to fail. The adaptive schemes are able to capture the correct solution. We also observe that the gradient boosted sampler \adaptiveg{} is more aggressive in localizing the points and shows the best prediction errors of less than 5\% for TC1 and TC2.}
\label{fig:poisson}
\end{figure}

\paragraph{Observations}
We start by examining the performance of the difference methods for the relatively small
number of collocation points of $n_c=1,000$ (which corresponds to 1.5\% of the $256 \times 256$ domain $\Omega$).
We report the errors $\mu_1$ and $\mu_2$ for the different schemes in \tref{tab:poisson}.
We also visualize the predicted solution for each method in~\fref{fig:poisson},
along with the location of the collocation points overlayed for the final model.
We can clearly see that the baseline model does not converge
(despite hyperparameter tuning) due to the optimizer stalls.
However, \resampling{} achieves significantly better errors, especially
for the smooth source function setup (\TCone{}).
However, for the sharp source experiment (\TCtwo{}), the resampling does not help and shows an error of about $50\%$. 

Overall, the adaptive schemes show much better (or comparable) prediction errors for both test-cases.
In particular, \adaptiver{} and \adaptiveg{} achieve about 2--5\% relative error ($\mu_1$) for both \TCone{} and \TCtwo{}.
The visual reconstruction shown in~\fref{fig:poisson} also shows the clearly improved prediction with the adaptive schemes
(last two columns), as compared to the baseline with/without resampling (first two columns).
Note that the \adaptiveg{} method assigns more collocation points around the sharp features.
This is due to the
fact that it uses the gradient information for its probability distribution, instead of the residual of the PDE which
is used in \adaptiver{} method.
To show the effect of the scheduler, we show an ablation study with the cosine-annealing scheduler in the appendix (see \fref{fig:ns}).

We then repeat the same experiment, but now varying the number of collocation points, $n_c$, from $500$ to $8K$,
and we report the relative error ($\mu_1$) in~\tref{tab:poisson_col}.
We observe a consistent trend, where the Baseline (PINN training without resampling) does not converge to a good solution,
whereas \adaptiveg{} consistently achieves up to an order of magnitude better error.
Also, note that the performance of the \resampling{} method significantly improves and becomes on par with \adaptiveg{} as we increase $n_c$.
This is somewhat expected since at large number of collocation points resampling will have the chance to sample
points near the areas with sharp features.
In \fref{fig:error_bars}, we show the errors for every test-case as a function of number of collocation points using 10 different random seed values to quantify the variance in the different methods. We observe that the adaptive schemes additionally show smaller variances across seeds, especially for the test-cases with sharp sources.

\begin{table}
		\caption{We report the relative and absolute errors for the predicted solution for the two test-cases \TCone{} and \TCtwo{} for the four different schemes. The vanilla PINN training (baseline) fails to converge to a good solution for small number of collocation points ($n_c$), despite
		tuning the hyperparameter.
		The \resampling{} method however, achieves significantly better results for the smooth testing case (\TCone{}), but
		incurs high errors for the second test case which exhibits sharper features (\TCtwo{}).
		To the contrary, the two adaptive schemes of \adaptiver{} and \adaptiveg{} consistently achieve better (or comparable) results than
		both baseline and \resampling{} for both test cases.
		\label{tab:poisson}
		}
		\centering
			\makebox[\textwidth]{\centering
				\begin{tabular}{cccLLL}
					\hline
					Test-case & Errors & Baseline & \resampling{} & \adaptiver{} & \adaptiveg{}  \\
					\hline
					\multirow{2}{*}{\centering\textit{TC1}} & $\mu_1$ & 5.34\nexp{1} & \textbf{1.61\nexpb{2}} & 2.56\nexp{2} & {1.80\nexp{2}} \\
	                & $\mu_2$ & 7.84\pexp{0} & 1.68\nexp{1} & 2.57\nexp{1} & \textbf{1.43\nexpb{1}} \\
	                \hline
	                \multirow{2}{*}{\centering\textit{TC2}} & $\mu_1$ & 7.09\nexp{1} & 4.83\nexp{1} & 6.03\nexp{2} & \textbf{4.08\nexpb{2}} \\
	                & $\mu_2$ & 15.0\pexp{1} & 9.51\pexp{0} & 7.20\nexp{1} & \textbf{5.04\nexpb{1}} \\
					\hline
			\end{tabular}}
	\end{table}

\begin{table}
		\caption{We report relative errors for the Baseline, \resampling{} and \adaptiveg{} schemes as a function of number of collocation points $n_c$ for five different values in $\{500,1000,2000,4000,8000\}$. 
		We observe that the \resampling{} method performs well at the larger collocation points regime and for the test-cases with smooth functions. \adaptiveg{} shows a consistent performance on-par or better than the \resampling{} across all $n_c$ values.
		\label{tab:poisson_col}
		}
		\centering
			\makebox[\textwidth]{\centering
				\begin{tabular}{ccccccc}
					\hline
					Test-case & Method & $n_c=500$ & $n_c=1000$ & $n_c=2000$ & $n_c=4000$ & $n_c=8000$ \\
					\hline
					& Baseline & 4.41\nexp{1} & 5.34\nexp{1} & 2.73\nexp{2} & 4.70\nexp{2} & 5.39\nexp{1} \\
					& \resampling{} & 2.53\nexp{2} & \textbf{1.61\nexpb{2}} & \textbf{2.14\nexpb{2}} & \textbf{2.10\nexpb{2}} & 1.98\nexp{2} \\
					\gd \multirow{-3}{*}{\centering\textit{TC1}}  & \adaptiveg{}& \textbf{1.94\nexpb{2}} & 1.80\nexp{2} & 2.17\nexp{2} & 2.59\nexp{2} & \textbf{1.79\nexp{2}} \\
					\hline
					& Baseline & 7.64\nexp{1} & 7.09\nexp{1} & 7.34\nexp{1} & 6.93\nexp{1}& 5.44\nexp{1} \\
					& \resampling{} & 3.85\nexp{1} & 4.83\nexp{1} & 6.74\nexp{2} & 4.68\nexp{2} & 5.85\nexp{2} \\
					\gd \multirow{-3}{*}{\centering\textit{TC2}} & \adaptiveg{} & \textbf{4.86\nexpb{2}} & \textbf{4.08\nexpb{2}} & \textbf{4.43\nexpb{2}} & \textbf{3.55\nexpb{2}} & \textbf{3.91\nexpb{2}} \\
					\hline
			\end{tabular}}
	\end{table}

\subsection{2D diffusion-advection equation}\label{sec:poisadv}
\begin{figure}
\begin{subfigure}{0.5\textwidth}
  \centering
  \includegraphics[width=\linewidth]{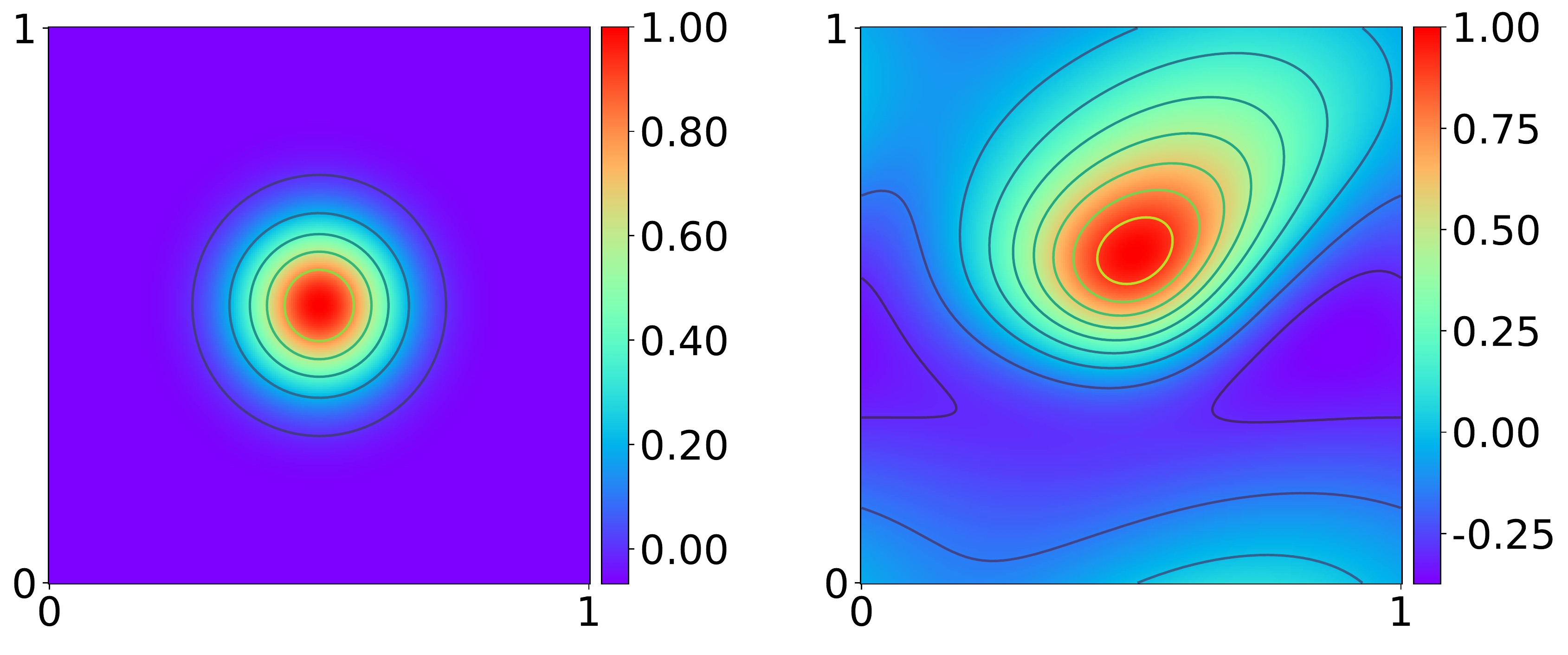}  
  \caption{$TC3$}
  \label{fig:poisadv-smooth}
\end{subfigure}%
\begin{subfigure}{0.5\textwidth}
  \centering
  \includegraphics[width=\linewidth]{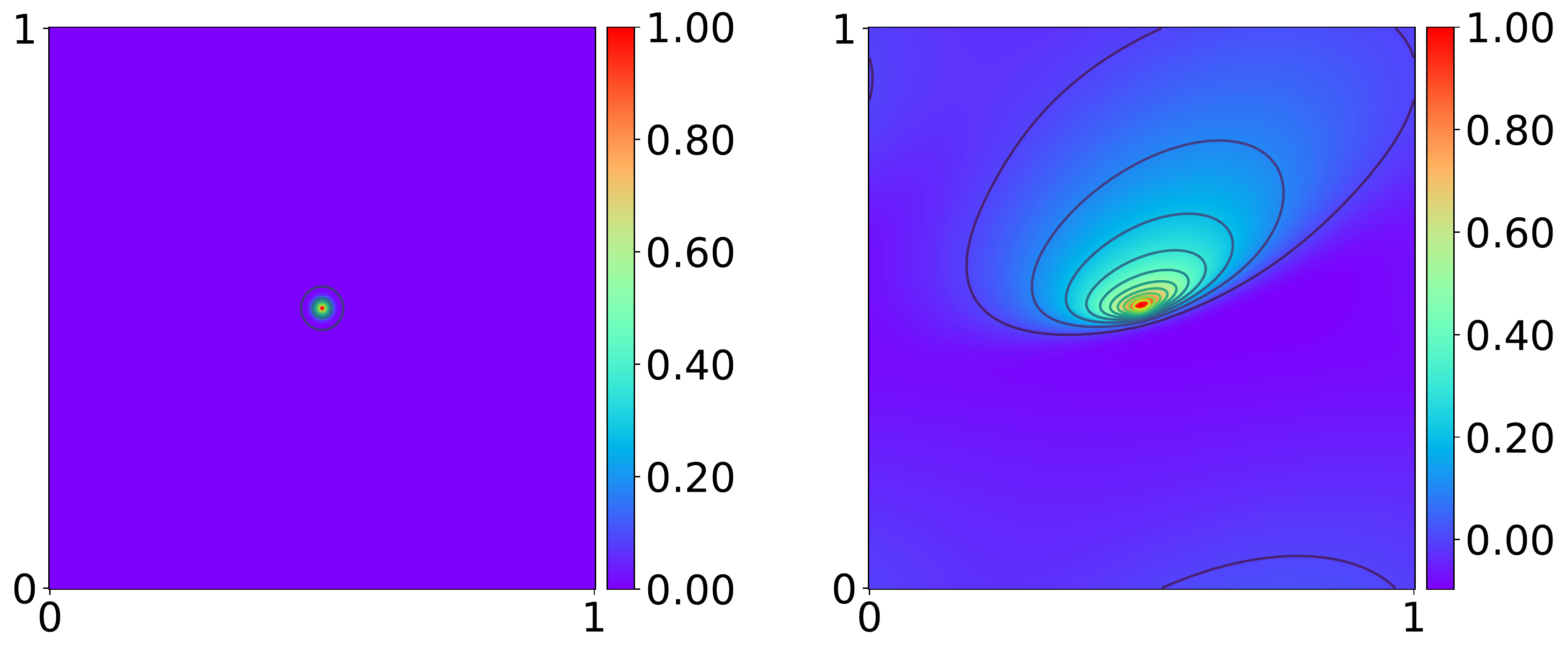}  
  \caption{$TC4$}
  \label{fig:poisadv-sharp}
\end{subfigure}
    \caption{We visualize the source and target solution pairs for TC3 (smooth source with $\sigma_f=0.1$, left) and TC4 (sharp source with $\sigma_f=0.01$, right) for the 2D diffusion-advection equation.}
    \label{fig:poisadv_systems}
\end{figure}

We next look at a steady-state 2D diffusion-advection equation with source function $f(\vect{x})$, diffusion tensor $\mat{K}$, and velocity vector $\vect{v}$:
\begin{equation}
    -\vect{v}\cdot \igrad u + \mat{K} \igrad u = f(\vect{x}), \quad \vect{x} \in \Omega.
    \label{eq:poisadv}
\end{equation}
For homogeneous diffusion tensors and velocity vectors, the solution of the advection-diffusion equation with doubly-periodic boundary conditions can also be computed using the fast Fourier transform on a discrete grid as:
\begin{equation}
\begin{split}
    \hat{u} &= F^{-1}\big(\frac{-1}{g_1 - g_2} F(f(\vect{x})) \big), \\
    g_1 &= -(k_x^2k_{11} + k_y^2k_{22} + 2k_xk_yk_{12}), \\
    g_2 &= ik_xv_1 + ik_yv_2,
\end{split}
\end{equation}
where $F$ is the Fourier transform, $i=\sqrt{-1}$,  $k_x, k_y$ are the frequencies in the Fourier domain, $k_{11}, k_{22}, k_{12}$ are the diagonal and off-diagonal coefficients of the diffusion tensor $\mat{K}$, and $v_1, v_2$ are the velocity components.
As before, we enforce the boundary conditions as Dirichlet boundary conditions using the true solution, and we consider a Gaussian function as the source with standard deviation $\sigma_f$. We use a diffusion tensor $k_{11}=1, k_{22}=8, k_{12}=4$ and velocity vector $v_1= 40, v_2=10$ to simulate sufficient advection. We consider two test-cases as before:
\begin{inparaenum}[(i)]
\item \TCthr{}: smooth source function with $\sigma_f = 0.1$
\item \TCfou{}: sharp source function with $\sigma_f = 0.01$.
\end{inparaenum}
We show the source functions and the target solutions in \fref{fig:poisadv_systems}. 

\begin{figure}
\begin{subfigure}{\textwidth}
  \centering
  \includegraphics[width=\linewidth]{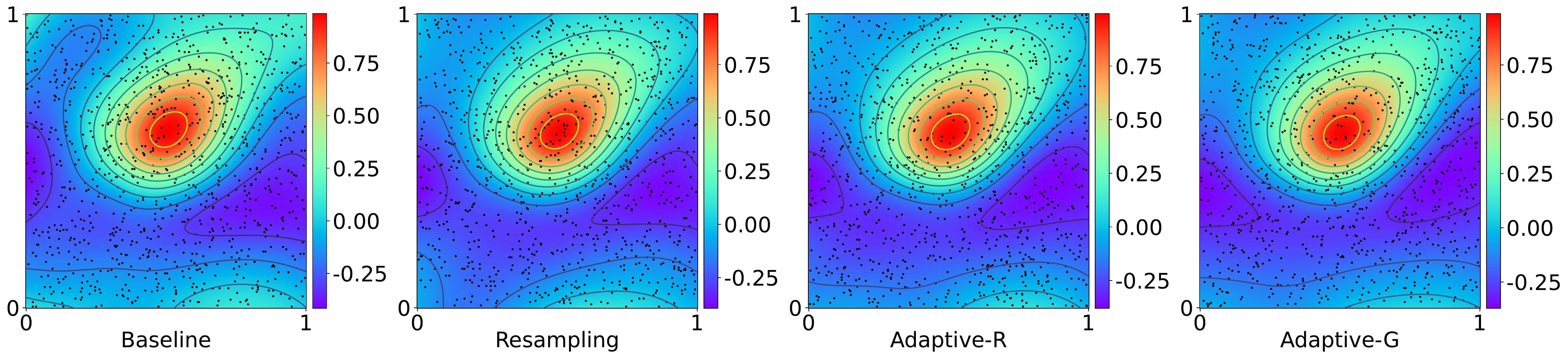}  
  \caption{$TC3$}
  \label{fig:poisadv-smooth-vis}
\end{subfigure}\\
\begin{subfigure}{\textwidth}
  \centering
  \includegraphics[width=\linewidth]{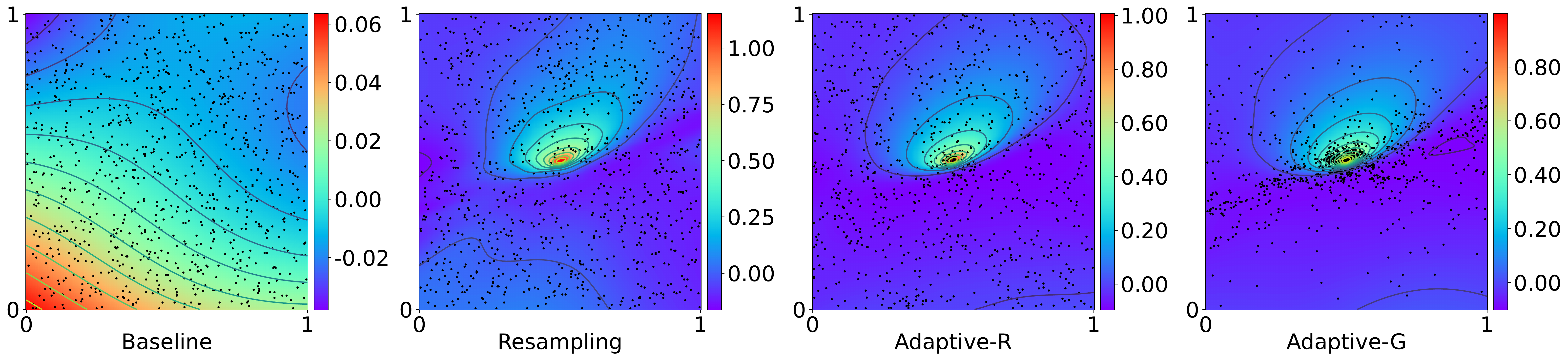}  
  \caption{$TC4$}
  \label{fig:poisadv-sharp-vis}
\end{subfigure}
\caption{We visualize the predicted solution at all testing points for both test-cases TC3 (top) and TC4 (bottom). We observe that resampling can help significantly for this system, but the adaptive schemes still show the least errors for both test-cases.}
\label{fig:poisadv}
\end{figure}

\paragraph{Observations}
We observe a very similar behaviour for the reconstruction errors as in the previous experiments.
As before, we start with $n_c=1,000$ collocation points, and we report the results in~\tref{tab:poisadv} and~\fref{fig:poisadv}.
Here, the baseline achieves slightly better performance for \TCthr{} (14\%) but completely fails (100\% error) for the \TCfou{} test case,
which includes a sharper forcing function.
However, the \resampling{} achieves better results for both cases.
Furthermore, the best performance is achieved by the adaptive methods.

Finally, in \tref{tab:poisadv_col} we report the relative errors for the baseline and adaptive schemes with various numbers of collocation points $n_c$. Similar to the Poisson system, larger values of $n_c$ show good performance, but the baselines underperform in the low data regime for sharp sources. 
The resampling achieves better errors, while the adaptive methods once again consistently achieve the best performance for both data regimes.

\begin{table}
		\caption{
		We report the relative and absolute errors for the predicted solution for the two test-cases \TCthr{}  and \TCfou{} for the four different schemes. As before, the Baseline fails to converge to a good solution.
		The \resampling{} method shows significant performance improvements. 
		The two adaptive schemes of \adaptiver{} and \adaptiveg{} consistently achieve better results than
		both baseline and \resampling{} for both test-cases.
		\label{tab:poisadv}
		}
		\centering
			\makebox[\textwidth]{\centering
				\begin{tabular}{cccLLL}
					\hline
					Test-case & Errors & Baseline & \resampling{}  & \adaptiver{} & \adaptiveg{}  \\
					\hline
					\multirow{2}{*}{\centering\textit{TC3}} & $\mu_1$ & 1.43\nexp{1} & 4.48\nexp{2} & \textbf{3.515\nexpb{2}} & {5.5\nexp{2}} \\
	                & $\mu_2$ & 3.23\nexp{1} & 1.63\nexp{1}  & 1.47\nexp{1} & \textbf{1.32\nexpb{1}} \\
	                \hline
	                \multirow{2}{*}{\centering\textit{TC4}} & $\mu_1$ & 1.06\pexp{0} & 4.10\nexp{1} &  \textbf{6.47\nexpb{2}} & {8.70\nexp{2}} \\
	                & $\mu_2$ & 5.23\pexp{0} & 1.83\pexp{0} & 3.28\nexp{1} & \textbf{3.29\nexpb{1}} \\
					\hline
			\end{tabular}}
	\end{table}

\begin{table}
		\caption{We report relative errors for the Baseline, \resampling{} and \adaptiveg{} schemes as a function of number of collocation points $n_c$ for five different values in $\{500,1000,2000,4000,8000\}$. As before, the \resampling{} method performs well at the larger collocation points regime and \adaptiveg{} shows a consistent performance across all $n_c$ values.
		\label{tab:poisadv_col}
		}
		\centering
			\makebox[\textwidth]{\centering
				\begin{tabular}{ccccccc}
					\hline
					Test-case & Method & $n_c=500$ & $n_c=1000$ & $n_c=2000$ & $n_c=4000$ & $n_c=8000$ \\
					\toprule
					& Baseline & 7.23\nexp{1} & 1.43\nexp{1} & 7.18\nexp{1} & 6.99\nexp{1} & 7.04\nexp{1} \\
					& \resampling{} & 4.69\nexp{2} & \textbf{4.48\nexpb{2}} & 5.10\nexp{2} & 5.83\nexp{2} & \textbf{3.82\nexpb{2}} \\
					\gd \multirow{-3}{*}{\centering\textit{TC3}} & \adaptiveg{} & \textbf{3.54\nexpb{2}} & 5.5\nexp{2} & \textbf{4\nexpb{2}} & \textbf{4.56\nexpb{2}} & 4.06\nexp{2} \\
					\midrule
					& Baseline & 1.094\pexp{0} & 1.07\pexp{0} & 1.09\pexp{0} & 1.04\pexp{0} & 9.91\nexp{1} \\
					& \resampling{} & 2.84\nexp{1} & 4.1\nexp{1} & \textbf{7.06\nexpb{2}} & 1.05\nexp{1} & 6.69\nexp{2} \\
					\gd \multirow{-3}{*}{\centering\textit{TC4}} & \adaptiveg{} & \textbf{1.13\nexpb{1}} & \textbf{8.70\nexpb{2}} & 7.47\nexp{2} & \textbf{5.69\nexpb{2}} & \textbf{5.03\nexpb{2}} \\
					\bottomrule
			\end{tabular}}
	\end{table}

\section{Conclusions}
We studied the impact of the location of the collocation points in training SciML models, focusing on the popular class of PINN models, and we showed that the vanilla PINN strategy of keeping the collocation points fixed throughout training often results in suboptimal solutions.
This is particularly the case for PDE systems with sharp (or very localized) features.
We showed that a simple strategy of resampling collocation points during optimization stalls can significantly improve the reconstruction error, especially for moderately large number of collocation points.
We also proposed adaptive collocation schemes to obtain a better allocation of the collocation points.
This is done by constructing 
a probability distribution derived from either the PDE residual (\adaptiver{}) or its gradient w.r.t. the input (\adaptiveg{}).
We found that by progressively incorporating the adaptive schemes, we can achieve up to an order of magnitude
better solutions, as compared to the baseline, especially for the regime of a small number of collocation points and with problems that exhibit sharp (or very localized) features.
Some limitations of this current work include the following: 
we did not change the NN architecture (it was fixed as a feed forward NN) or tune hyperparameters relating to the architecture (which can be a significant factor in any analysis); and
we only focused on 2D spatial systems (it is known that time-dependent or 3D or higher-dimensional systems can show different behaviours and may benefit from different kinds of adaptivity in space and time). 
We leave these directions to future work.
However, we expect that techniques such as those we used here that aim to combine in a more principled way domain-driven scientific methods and data-driven ML methods will help in these cases as well.

\section{Acknowledgements}
This research used resources of the National Energy Research Scientific Computing Center (NERSC), a U.S. Department of Energy Office of Science User Facility located at Lawrence Berkeley National Laboratory.
MWM would like to acknowledge the DOE, NSF, and ONR for providing partial support of this work.
AG was supported through funding from Samsung SAIT.

\bibliographystyle{plain}
\bibliography{references}

\begin{thebibliography}{10}

\bibitem{github}
\url{https://github.com/ShashankSubramanian/adaptive-selfsupervision-pinns}.

\bibitem{bischof2021multi}
Rafael Bischof and Michael Kraus.
\newblock Multi-objective loss balancing for physics-informed deep learning.
\newblock {\em arXiv preprint arXiv:2110.09813}, 2021.

\bibitem{boyd2004convex}
Stephen Boyd and Lieven Vandenberghe.
\newblock {\em Convex optimization}.
\newblock Cambridge university press, 2004.

\bibitem{brunton2020machine}
Steven~L Brunton, Bernd~R Noack, and Petros Koumoutsakos.
\newblock Machine learning for fluid mechanics.
\newblock {\em Annual Review of Fluid Mechanics}, 52:477--508, 2020.

\bibitem{chen2020physics}
Yuyao Chen, Lu~Lu, George~Em Karniadakis, and Luca Dal~Negro.
\newblock Physics-informed neural networks for inverse problems in nano-optics
  and metamaterials.
\newblock {\em Optics express}, 28(8):11618--11633, 2020.

\bibitem{EdwCACM22}
C.~Edwards.
\newblock Neural networks learn to speed up simulations.
\newblock {\em Communications of the ACM}, 65(5):27--29, 2022.

\bibitem{geneva2020modeling}
Nicholas Geneva and Nicholas Zabaras.
\newblock Modeling the dynamics of pde systems with physics-constrained deep
  auto-regressive networks.
\newblock {\em Journal of Computational Physics}, 403:109056, 2020.

\bibitem{hanna2021residual}
John Hanna, Jose~V Aguado, Sebastien Comas-Cardona, Ramzi Askri, and Domenico
  Borzacchiello.
\newblock Residual-based adaptivity for two-phase flow simulation in porous
  media using physics-informed neural networks.
\newblock {\em arXiv preprint arXiv:2109.14290}, 2021.

\bibitem{jin2021nsfnets}
Xiaowei Jin, Shengze Cai, Hui Li, and George~Em Karniadakis.
\newblock Nsfnets (navier-stokes flow nets): Physics-informed neural networks
  for the incompressible navier-stokes equations.
\newblock {\em Journal of Computational Physics}, 426:109951, 2021.

\bibitem{karniadakis2021review}
George~Em Karniadakis, Ioannis~G Kevrekidis, Lu~Lu, Paris Perdikaris, Sifan
  Wang, and Liu Yang.
\newblock Physics-informed machine learning.
\newblock {\em Nature Reviews Physics}, 3(6):422--440, 2021.

\bibitem{krishnapriyan2021characterizing}
Aditi Krishnapriyan, Amir Gholami, Shandian Zhe, Robert Kirby, and Michael~W
  Mahoney.
\newblock Characterizing possible failure modes in physics-informed neural
  networks.
\newblock {\em Advances in Neural Information Processing Systems}, 34, 2021.

\bibitem{lagaris1998artificial}
Isaac~E Lagaris, Aristidis Likas, and Dimitrios~I Fotiadis.
\newblock Artificial neural networks for solving ordinary and partial
  differential equations.
\newblock {\em IEEE transactions on neural networks}, 9(5):987--1000, 1998.

\bibitem{liu2021dual}
Dehao Liu and Yan Wang.
\newblock A dual-dimer method for training physics-constrained neural networks
  with minimax architecture.
\newblock {\em Neural Networks}, 136:112--125, 2021.

\bibitem{lu2021deepxde}
Lu~Lu, Xuhui Meng, Zhiping Mao, and George~Em Karniadakis.
\newblock Deepxde: A deep learning library for solving differential equations.
\newblock {\em SIAM Review}, 63(1):208--228, 2021.

\bibitem{mcclenny2020self}
Levi McClenny and Ulisses Braga-Neto.
\newblock Self-adaptive physics-informed neural networks using a soft attention
  mechanism.
\newblock {\em arXiv preprint arXiv:2009.04544}, 2020.

\bibitem{rackauckas2020universal}
Christopher Rackauckas, Yingbo Ma, Julius Martensen, Collin Warner, Kirill
  Zubov, Rohit Supekar, Dominic Skinner, Ali Ramadhan, and Alan Edelman.
\newblock Universal differential equations for scientific machine learning.
\newblock {\em arXiv preprint arXiv:2001.04385}, 2020.

\bibitem{raissi2019physics}
Maziar Raissi, Paris Perdikaris, and George~E Karniadakis.
\newblock Physics-informed neural networks: A deep learning framework for
  solving forward and inverse problems involving nonlinear partial differential
  equations.
\newblock {\em Journal of Computational Physics}, 378:686--707, 2019.

\bibitem{raissi2020hidden}
Maziar Raissi, Alireza Yazdani, and George~Em Karniadakis.
\newblock Hidden fluid mechanics: Learning velocity and pressure fields from
  flow visualizations.
\newblock {\em Science}, 367(6481):1026--1030, 2020.

\bibitem{ramabathiran2021spinn}
Amuthan~A Ramabathiran and Prabhu Ramachandran.
\newblock Spinn: Sparse, physics-based, and partially interpretable neural
  networks for pdes.
\newblock {\em Journal of Computational Physics}, 445:110600, 2021.

\bibitem{sahli2020physics}
Francisco Sahli~Costabal, Yibo Yang, Paris Perdikaris, Daniel~E Hurtado, and
  Ellen Kuhl.
\newblock Physics-informed neural networks for cardiac activation mapping.
\newblock {\em Frontiers in Physics}, 8:42, 2020.

\bibitem{sirignano2018dgm}
Justin Sirignano and Konstantinos Spiliopoulos.
\newblock Dgm: A deep learning algorithm for solving partial differential
  equations.
\newblock {\em Journal of computational physics}, 375:1339--1364, 2018.

\bibitem{tang2021deep}
Kejun Tang, Xiaoliang Wan, and Chao Yang.
\newblock Das: A deep adaptive sampling method for solving partial differential
  equations.
\newblock {\em arXiv preprint arXiv:2112.14038}, 2021.

\bibitem{von2019informed}
Laura von Rueden, Sebastian Mayer, Katharina Beckh, Bogdan Georgiev, Sven
  Giesselbach, Raoul Heese, Birgit Kirsch, Julius Pfrommer, Annika Pick,
  Rajkumar Ramamurthy, et~al.
\newblock Informed machine learning--a taxonomy and survey of integrating
  knowledge into learning systems.
\newblock {\em arXiv preprint arXiv:1903.12394}, 2019.

\bibitem{wang2022respecting}
Sifan Wang, Shyam Sankaran, and Paris Perdikaris.
\newblock Respecting causality is all you need for training physics-informed
  neural networks.
\newblock {\em arXiv preprint arXiv:2203.07404}, 2022.

\bibitem{wang2020understanding}
Sifan Wang, Yujun Teng, and Paris Perdikaris.
\newblock Understanding and mitigating gradient pathologies in physics-informed
  neural networks.
\newblock {\em arXiv preprint arXiv:2001.04536}, 2020.

\bibitem{wang2020eigenvector}
Sifan Wang, Hanwen Wang, and Paris Perdikaris.
\newblock On the eigenvector bias of fourier feature networks: From regression
  to solving multi-scale pdes with physics-informed neural networks.
\newblock {\em arXiv preprint arXiv:2012.10047}, 2020.

\bibitem{wang2020and}
Sifan Wang, Xinling Yu, and Paris Perdikaris.
\newblock When and why pinns fail to train: A neural tangent kernel
  perspective.
\newblock {\em arXiv preprint arXiv:2007.14527}, 2020.

\bibitem{willard2020integrating}
Jared Willard, Xiaowei Jia, Shaoming Xu, Michael Steinbach, and Vipin Kumar.
\newblock Integrating physics-based modeling with machine learning: A survey.
\newblock {\em arXiv preprint arXiv:2003.04919}, 2020.

\bibitem{xiang2021self}
Zixue Xiang, Wei Peng, Xiaohu Zheng, Xiaoyu Zhao, and Wen Yao.
\newblock Self-adaptive loss balanced physics-informed neural networks for the
  incompressible navier-stokes equations.
\newblock {\em arXiv preprint arXiv:2104.06217}, 2021.

\bibitem{zhu2019physics}
Yinhao Zhu, Nicholas Zabaras, Phaedon-Stelios Koutsourelakis, and Paris
  Perdikaris.
\newblock Physics-constrained deep learning for high-dimensional surrogate
  modeling and uncertainty quantification without labeled data.
\newblock {\em Journal of Computational Physics}, 394:56--81, 2019.

\end{thebibliography}

\clearpage
\counterwithin{figure}{section}
\counterwithin{table}{section}
\appendix
\section{Additional results}
\begin{figure}
\begin{subfigure}{0.5\textwidth}
  \centering
  \includegraphics[width=\linewidth]{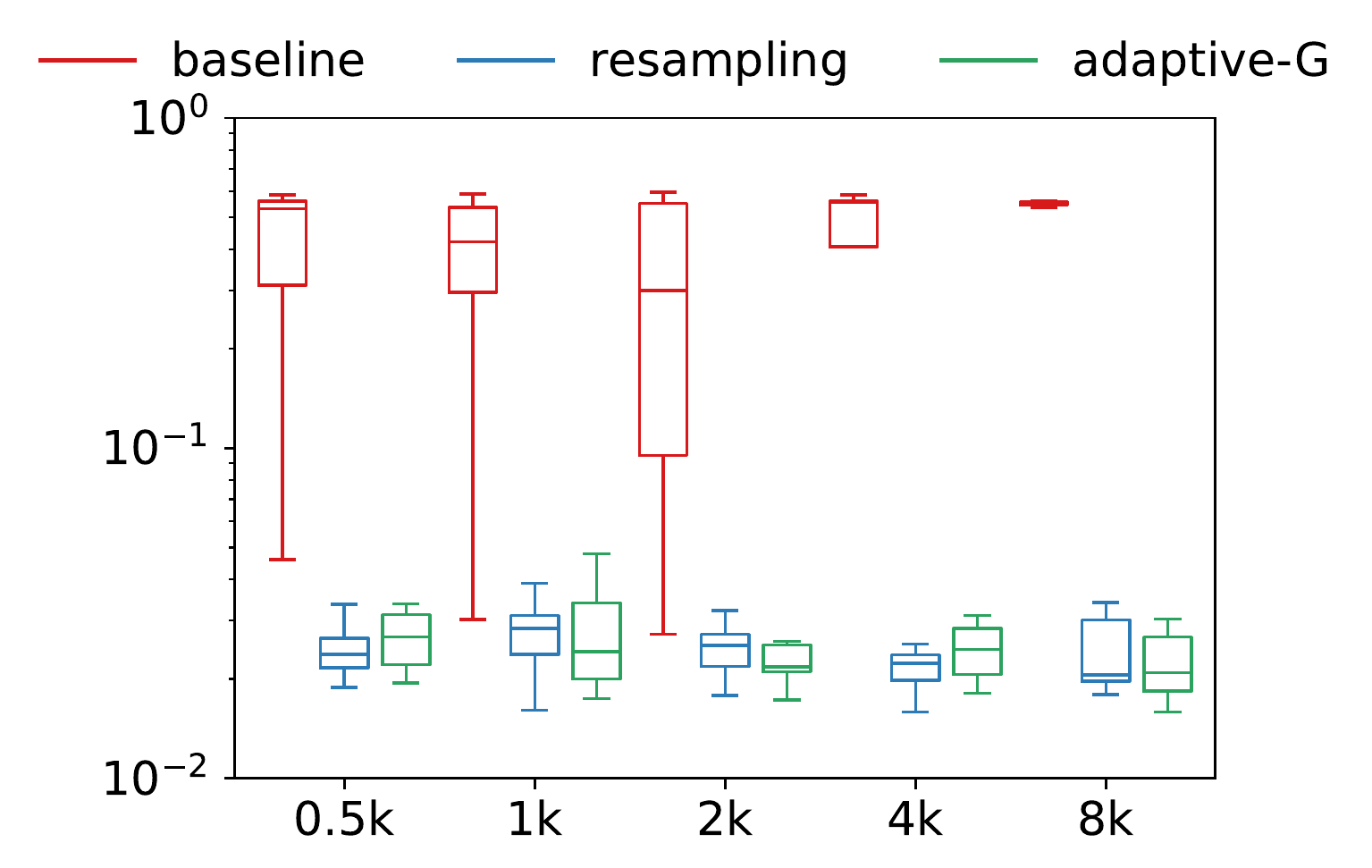}  
  \caption{$TC1$}
  \label{fig:err_tc1}
\end{subfigure}
\begin{subfigure}{0.5\textwidth}
  \centering
  \includegraphics[width=\linewidth]{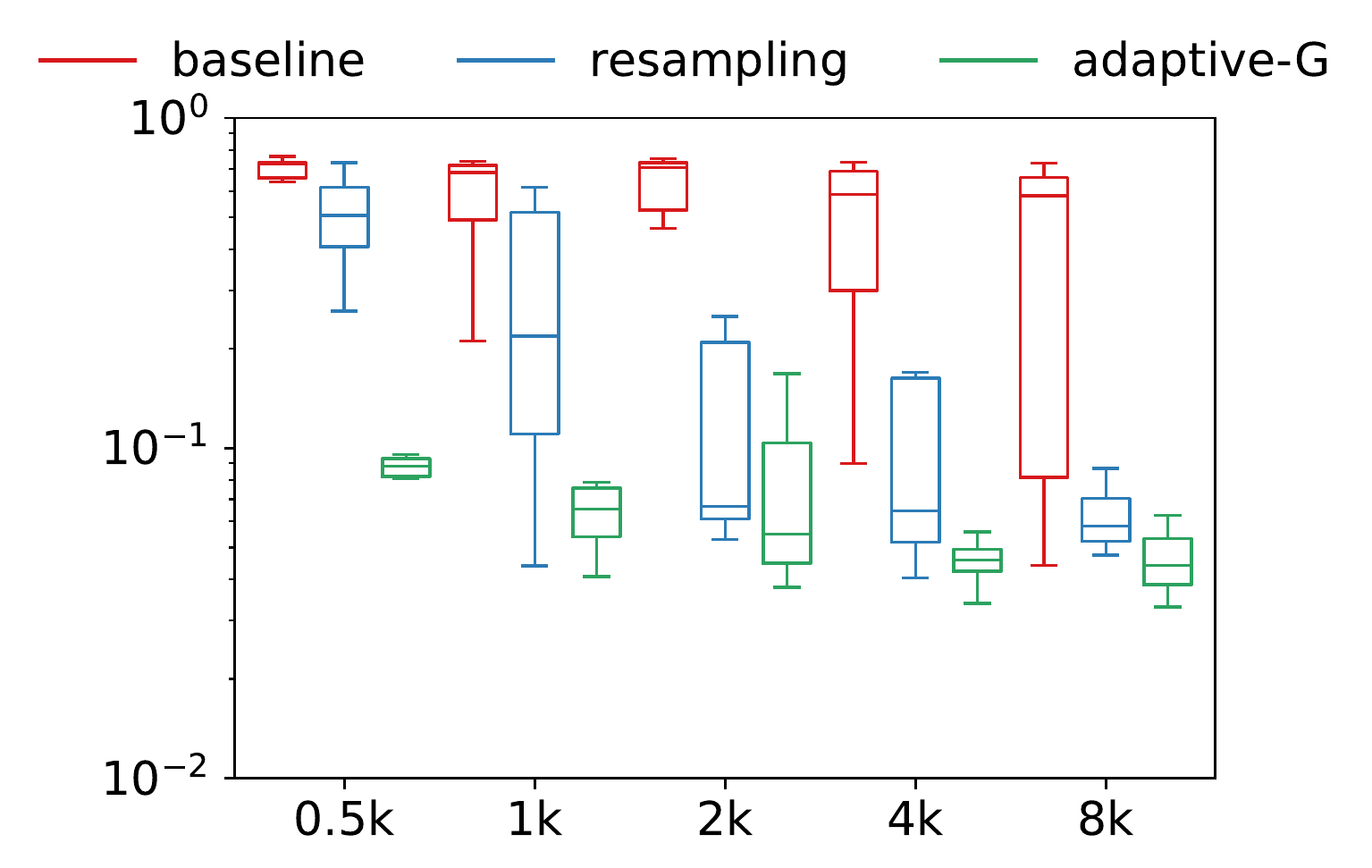}  
  \caption{$TC2$}
  \label{fig:err_tc2}
\end{subfigure}\\
\begin{subfigure}{0.5\textwidth}
  \centering
  \includegraphics[width=\linewidth]{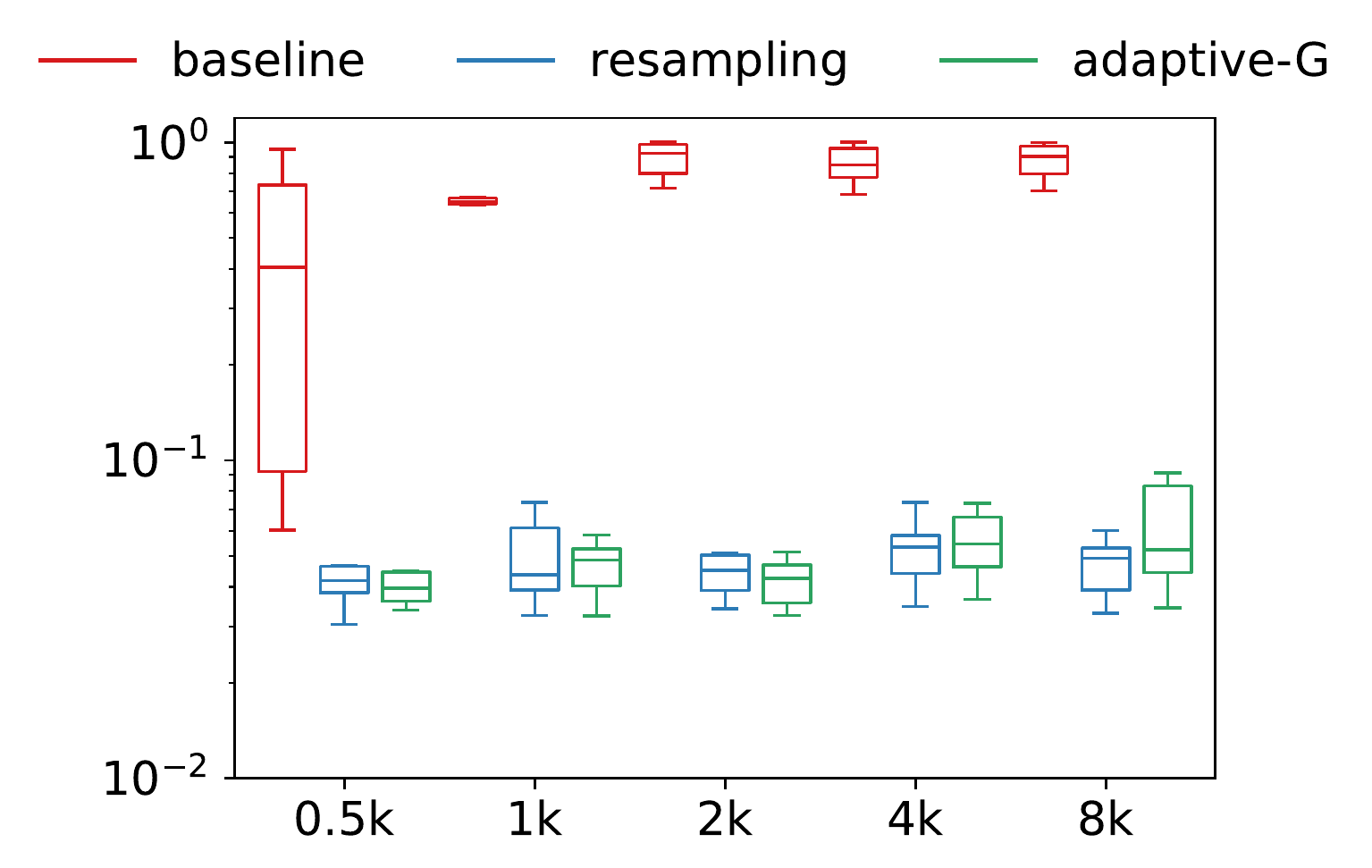}  
  \caption{$TC3$}
  \label{fig:err_tc3}
\end{subfigure}
\begin{subfigure}{0.5\textwidth}
  \centering
  \includegraphics[width=\linewidth]{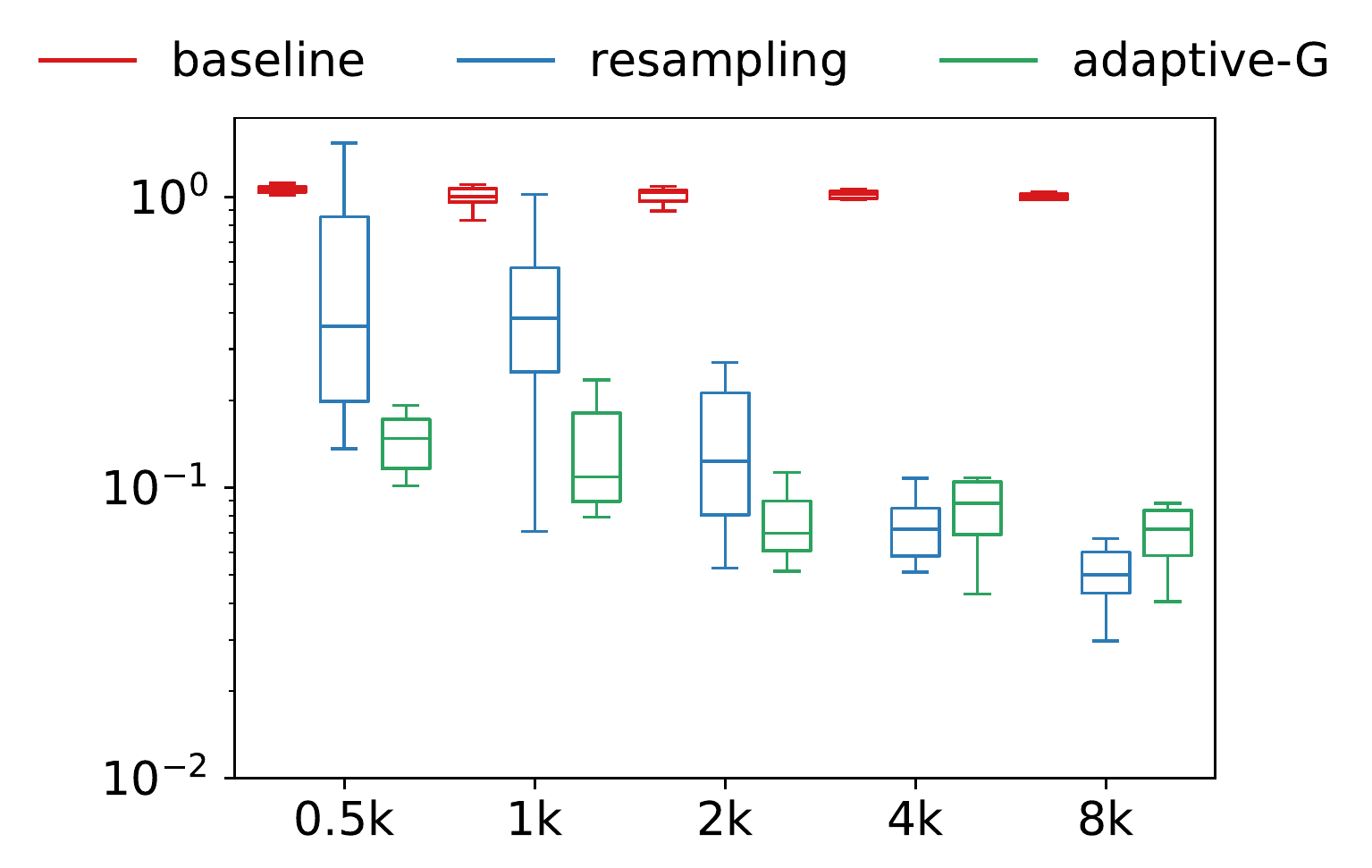}  
  \caption{$TC4$}
  \label{fig:err_tc4}
\end{subfigure}
\caption{Summary of testing errors for baseline, \resampling{} and \adaptiveg{} schemes as a function of different collocation points for all the test-cases to show the variance in error across 10 preset random seeds.}
\label{fig:error_bars}
\end{figure}

\paragraph{Ablation with different random seeds}
We repeat all experiments with 10 different random seeds to quantify the variance in our metrics. We report the testing error and the variance in \fref{fig:error_bars}. We observe that the adaptive schemes consistently show the best (or comparable) performance; and that they are superior to the resampling scheme in the low collocation points regime as well as for \TCtwo{} and \TCfou{} (systems with sharp features). Further, the variance in the error for the adaptive schemes is much smaller than the resampling for these two test-cases. For the smooth systems, both schemes show comparable performance and spread.

\begin{figure}
\begin{subfigure}{\textwidth}
  \centering
  \includegraphics[width=.8\linewidth]{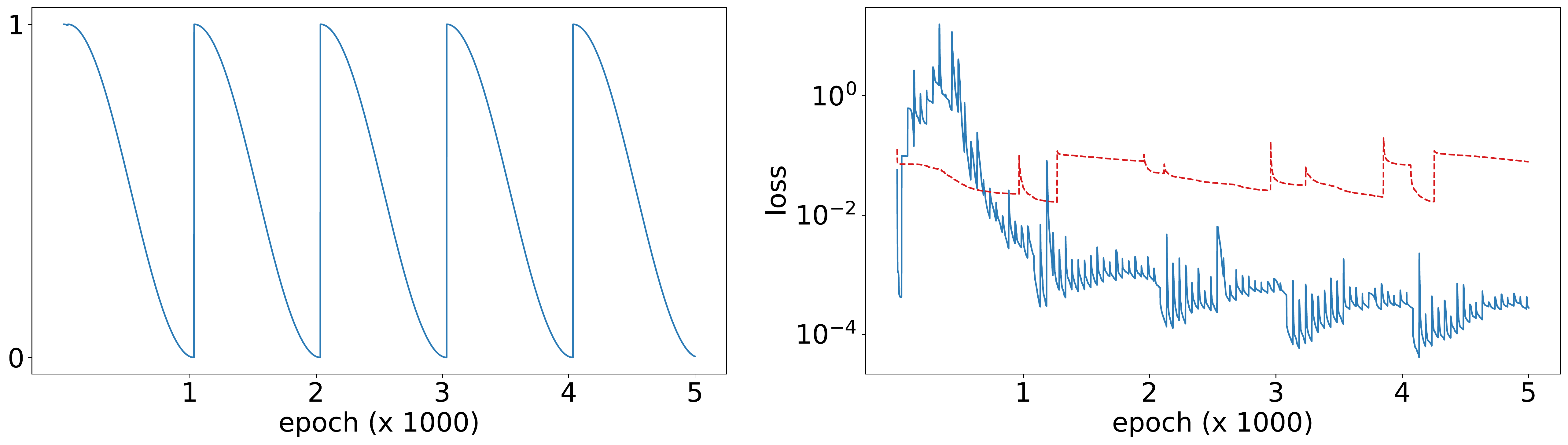}  
  \caption{$n_c = 500$}
\end{subfigure}\\
\begin{subfigure}{\textwidth}
  \centering
  \includegraphics[width=.8\linewidth]{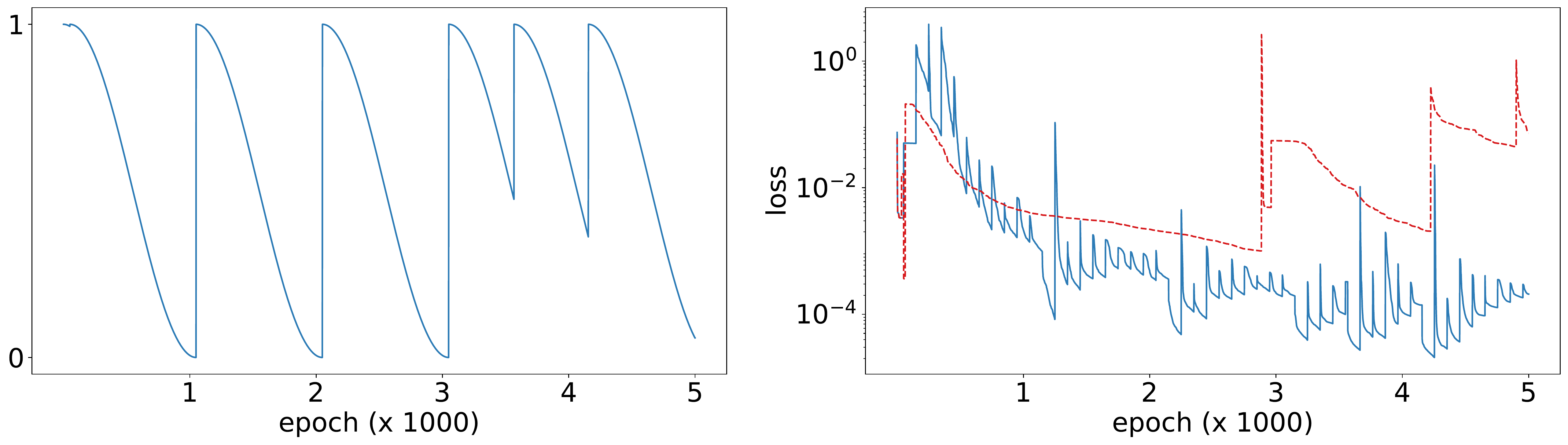}  
  \caption{$n_c = 1000$}
\end{subfigure}\\
\begin{subfigure}{\textwidth}
  \centering
  \includegraphics[width=.8\linewidth]{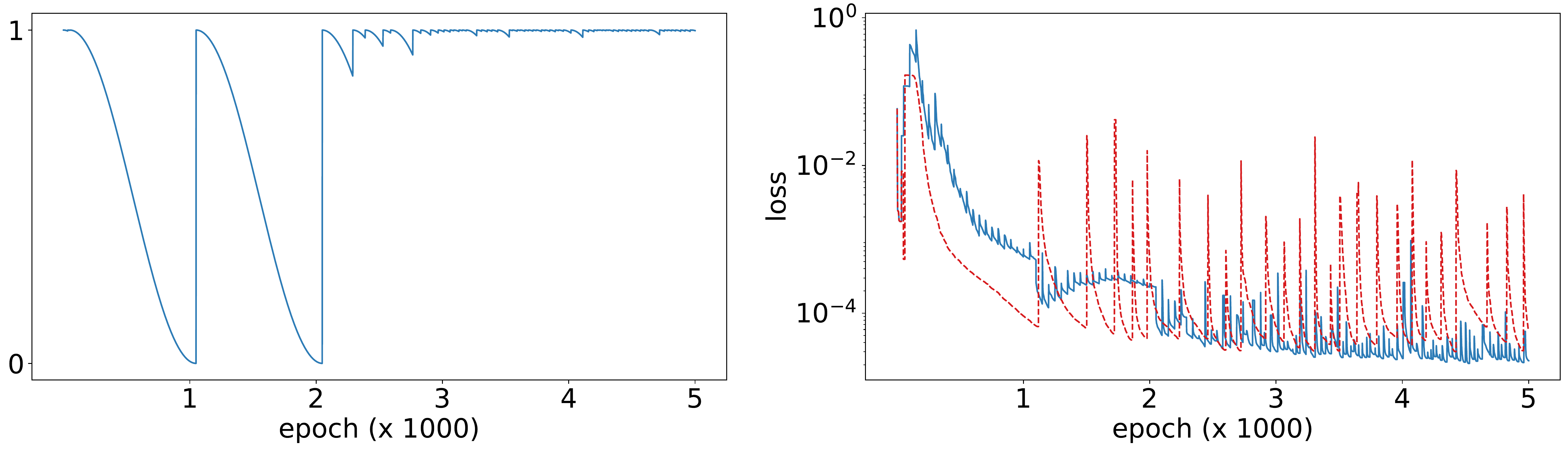}  
  \caption{$n_c = 2000$}
\end{subfigure}\\
\begin{subfigure}{\textwidth}
  \centering
  \includegraphics[width=.8\linewidth]{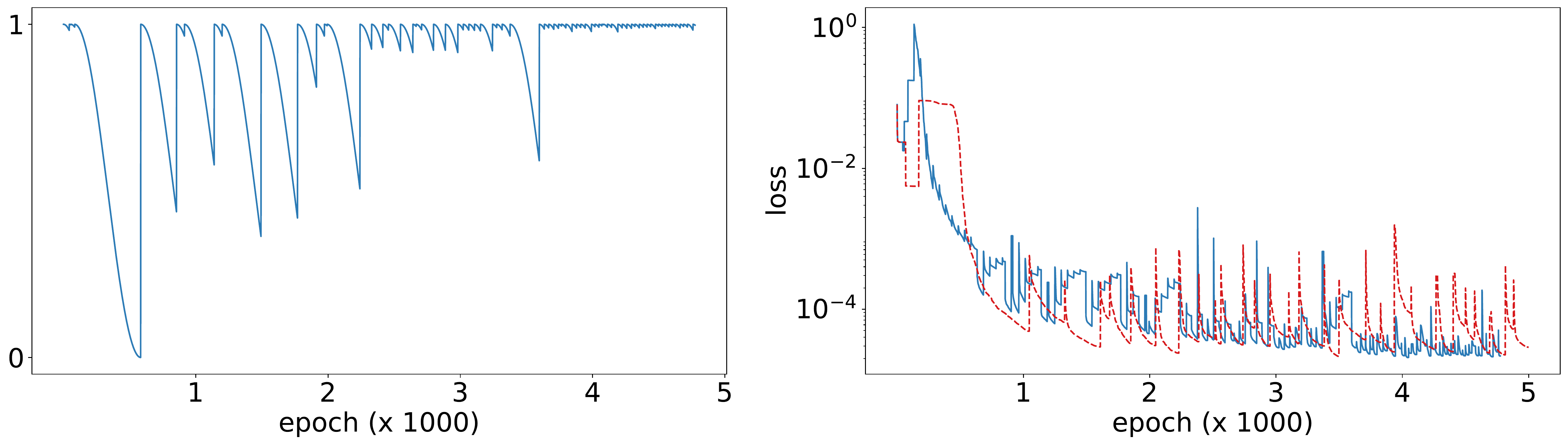}  
  \caption{$n_c = 4000$}
\end{subfigure}\\
\begin{subfigure}{\textwidth}
  \centering
  \includegraphics[width=.8\linewidth]{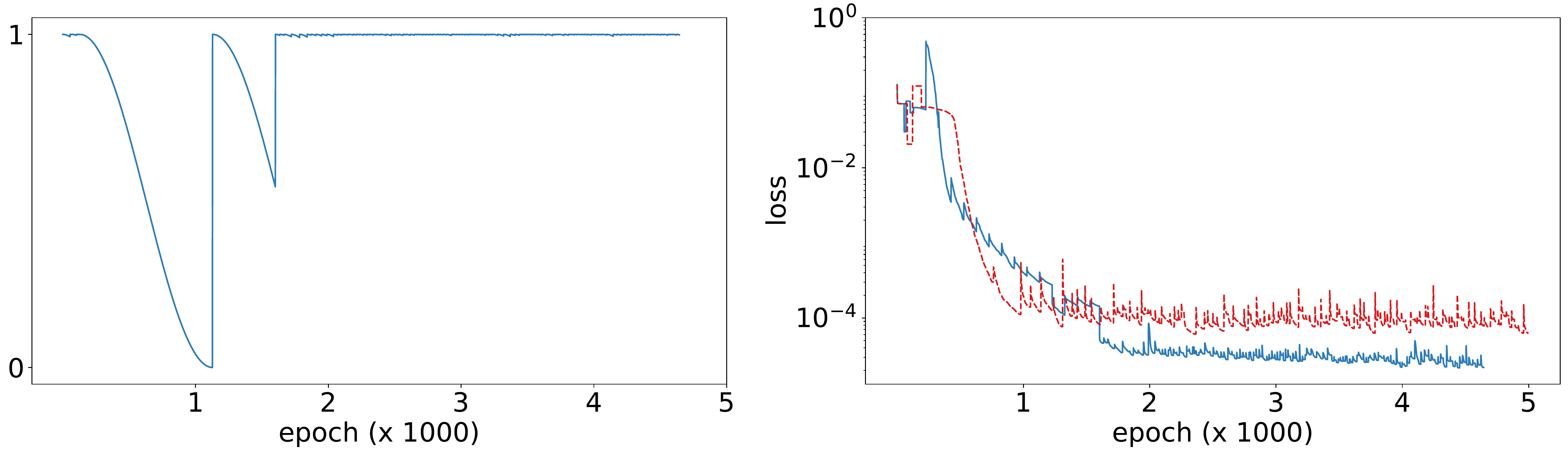}  
  \caption{$n_c = 8000$}
\end{subfigure}\\
\caption{We show the scheduler fraction $\eta$ (left; fraction of uniform sampling) and the training loss curves (right) for the \resampling{} (red) and \adaptiveg{} (blue) schemes for \TCtwo{} for different $n_c$ values.}
\label{fig:pois_loss}
\end{figure}

\begin{figure}
\begin{subfigure}{\textwidth}
  \centering
  \includegraphics[width=.8\linewidth]{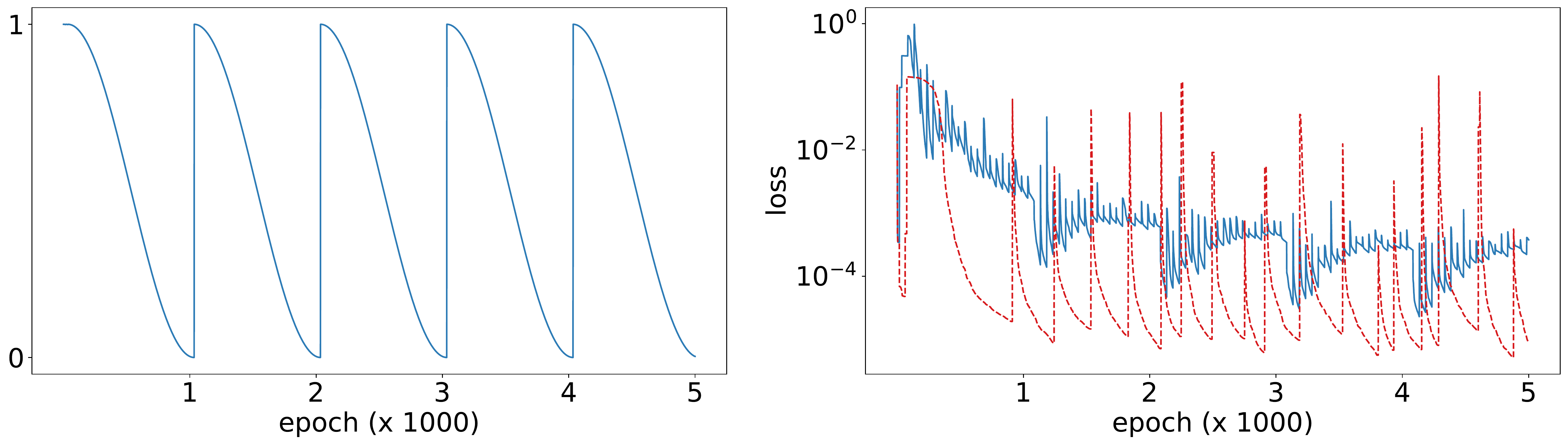}  
  \caption{$n_c = 500$}
\end{subfigure}\\
\begin{subfigure}{\textwidth}
  \centering
  \includegraphics[width=.8\linewidth]{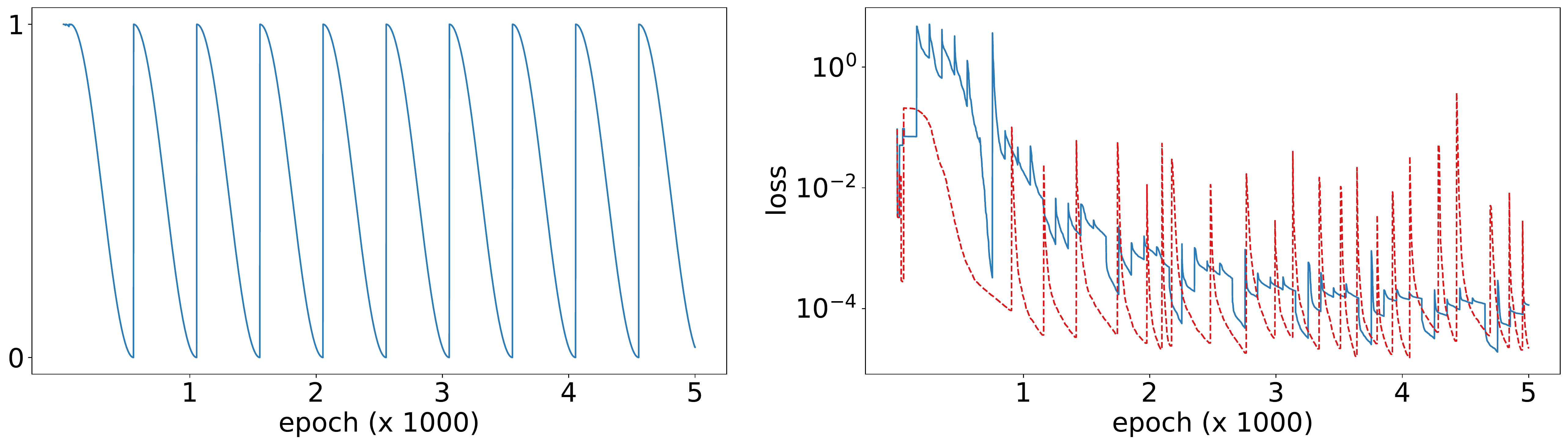}  
  \caption{$n_c = 1000$}
\end{subfigure}\\
\begin{subfigure}{\textwidth}
  \centering
  \includegraphics[width=.8\linewidth]{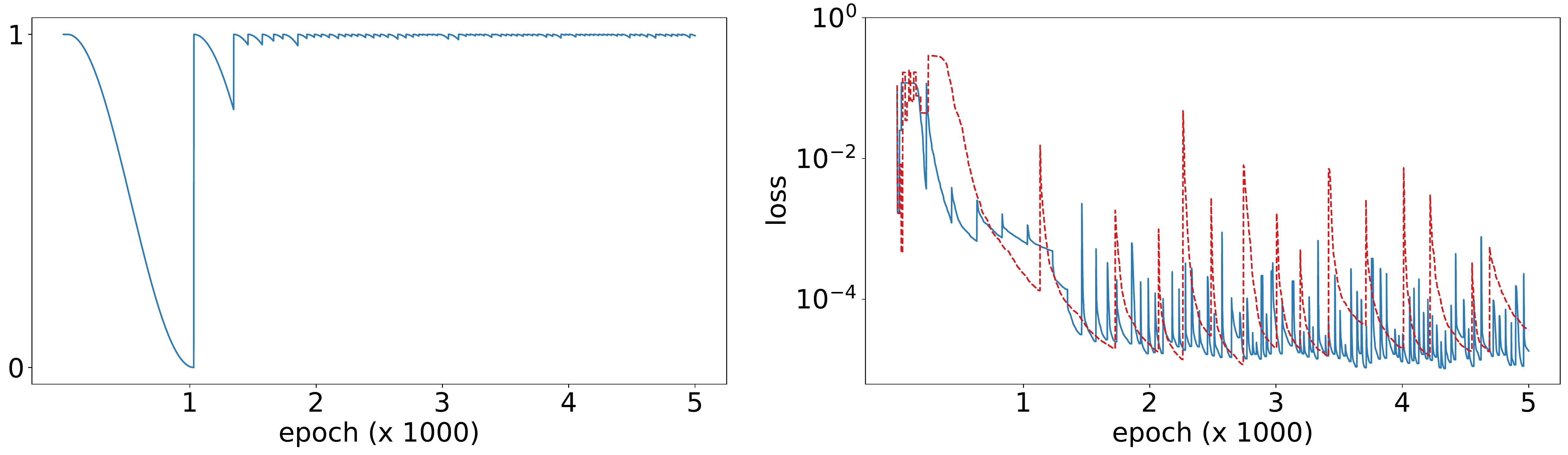}  
  \caption{$n_c = 2000$}
\end{subfigure}\\
\begin{subfigure}{\textwidth}
  \centering
  \includegraphics[width=.8\linewidth]{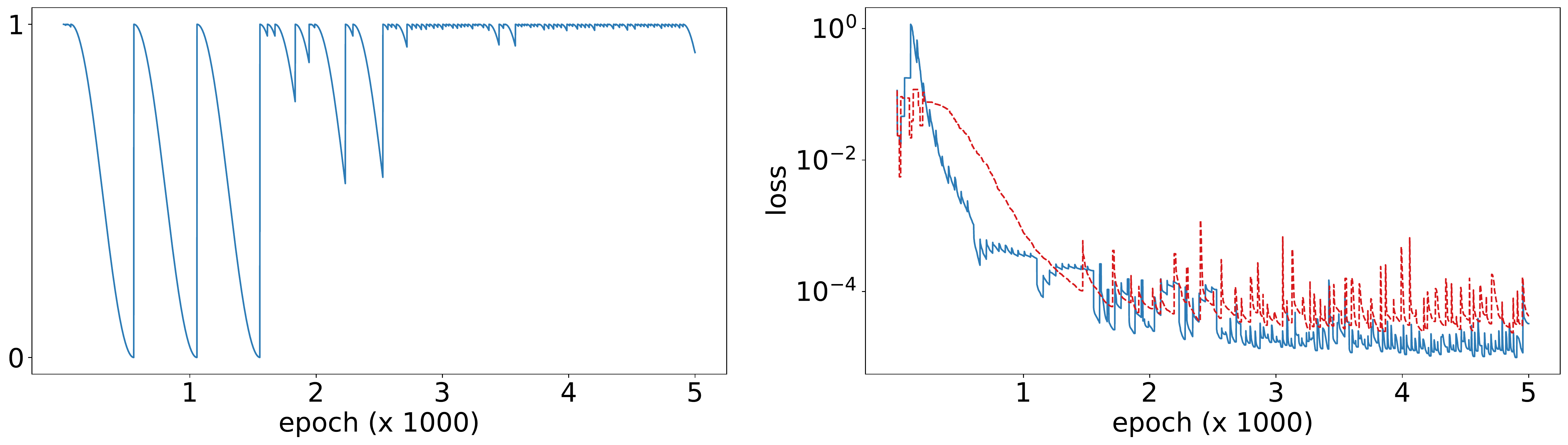}  
  \caption{$n_c = 4000$}
\end{subfigure}\\
\begin{subfigure}{\textwidth}
  \centering
  \includegraphics[width=.8\linewidth]{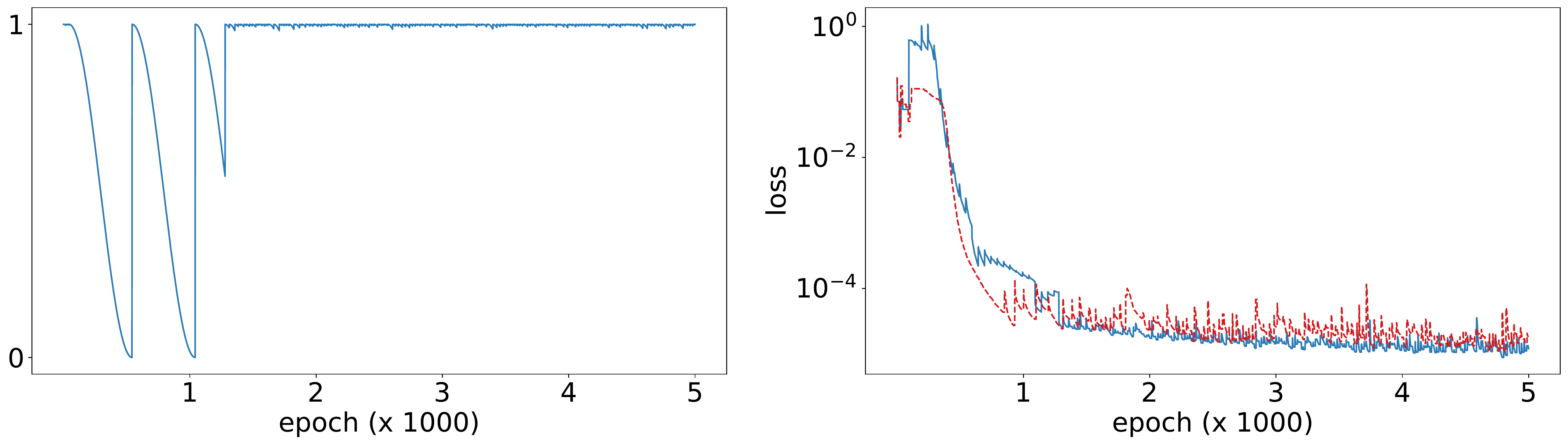}  
  \caption{$n_c = 8000$}
\end{subfigure}\\
\caption{We show the scheduler fraction $\eta$ (left; fraction of uniform sampling) and the training loss curves (right) for the \resampling{} (red) and \adaptiveg{} (blue) schemes for \TCfou{} for different $n_c$ values.}
\label{fig:poisadv_loss}
\end{figure}

\paragraph{Ablation with loss and schedules for different systems}
We show the annealing schedule and the training loss curves for \TCtwo{} and \TCfou{} in \fref{fig:pois_loss} and \fref{fig:poisadv_loss} as a function of different $n_c$ values. We observe that the optimization difficulties increase with larger $n_c$, leading the scheduler to restart more often, preferring a uniform sampling. The training loss curves for the \resampling{} and \adaptiveg{} schemes also become more similar as $n_c$ increases. 

\begin{figure}
    \centering
    \includegraphics[width=0.75\linewidth]{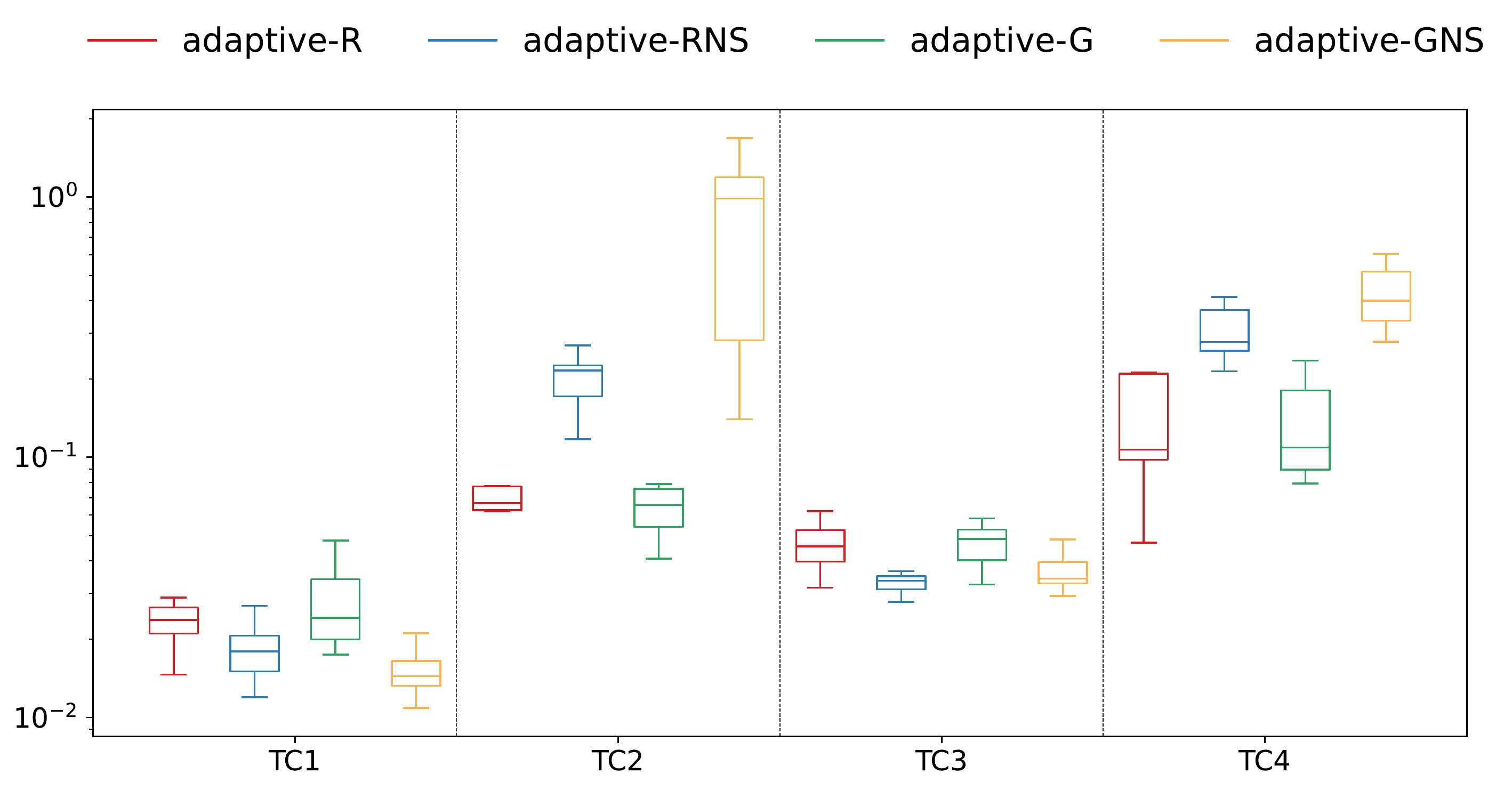}
\caption{We report the relative errors for \adaptiverns{} and \adaptivegns{}, which are the adaptive schemes with no scheduler for the PDE residual and loss gradient respectively and compare them to their counterparts with the cosine-annealing schedule \adaptiver{} and \adaptiveg{} across 10 random seeds to additionally show the variance in the runs.}
\label{fig:ns}
\end{figure}

\paragraph{Ablation with scheduler}
We run the adaptive schemes without any scheduling to demonstrate the effect of the scheduler. Specifically, we run \adaptiverns{} and \adaptivegns{}, which are the adaptive schemes with no schedule (NS) for the PDE residual and loss gradient proxy, respectively. We report our results for $n_c=1000$ for all test-cases in \fref{fig:ns}. 
We observe that while the non-scheduled adaptive schemes are quite performant, they show higher errors and larger variability (across different random seeds) for the test-cases with sharp features. Specifically, for the NS schemes, \TCtwo{} and \TCfou{} have median errors of 20--30\%, while with the schedule the errors drop to 6--10\%. Hence, the advantage of a scheduled sampling is greater for systems with sharper features.

\end{document}